\documentclass[10pt,twocolumn,letterpaper]{article}

\usepackage{cvpr}              
\usepackage{multirow}
\usepackage[dvipsnames]{xcolor}

\usepackage{makecell}
\usepackage{algorithm}
\usepackage{algorithmic}
\usepackage[misc]{ifsym}

\usepackage{pifont}
\newcommand{\cmark}{\textcolor{black}{\ding{51}}}
\newcommand{\xmark}{\textcolor{black}{\ding{55}}}%

\definecolor{citeblue}{RGB}{48,111,186}
\usepackage[colorlinks,backref=true,linkcolor=black, citecolor=citeblue,urlcolor=purple]{hyperref}
\definecolor{cvprblue}{rgb}{0.21,0.49,0.74}

\def\paperID{2341} 
\def\confName{CVPR}
\def\confYear{2024}

\title{\raisebox{-0.6ex}{\protect\includegraphics[height=2.5\fontcharht\font`\B]{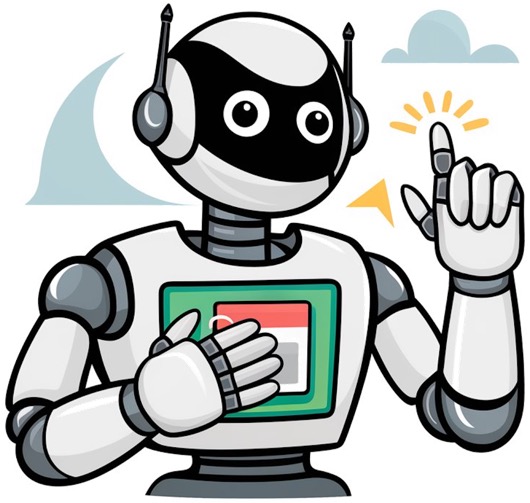}} \ CLOVA: A \underline{C}losed-\underline{LO}op \underline{V}isual \underline{A}ssistant with Tool Usage and Update}

\author{
Zhi Gao\textsuperscript{\rm1,2},
Yuntao Du\textsuperscript{\rm2},
Xintong Zhang\textsuperscript{\rm2,3},
Xiaojian Ma\textsuperscript{\rm2}, \\
Wenjuan Han\textsuperscript{\rm3},
Song-Chun Zhu\textsuperscript{\rm1,2,4},
Qing Li\textsuperscript{\rm2 \Letter} \\
 \small \textsuperscript{\rm 1}School of Intelligence Science and Technology, Peking University 
 \small \textsuperscript{\rm 2}State Key Laboratory of General Artificial Intelligence, BIGAI \\
 \small \textsuperscript{\rm 3}Beijing Jiaotong University 
 \small \textsuperscript{\rm 4}Department of Automation, Tsinghua University \\
 \small \href{https://clova-tool.github.io/}{https://clova-tool.github.io}
}

\begin{document}

\twocolumn[{%
\renewcommand\twocolumn[1][]{#1}%
\maketitle
\begin{center}
    \centering
    \captionsetup{type=figure}
    \vskip -0.35in
    \includegraphics[width=0.98\textwidth]{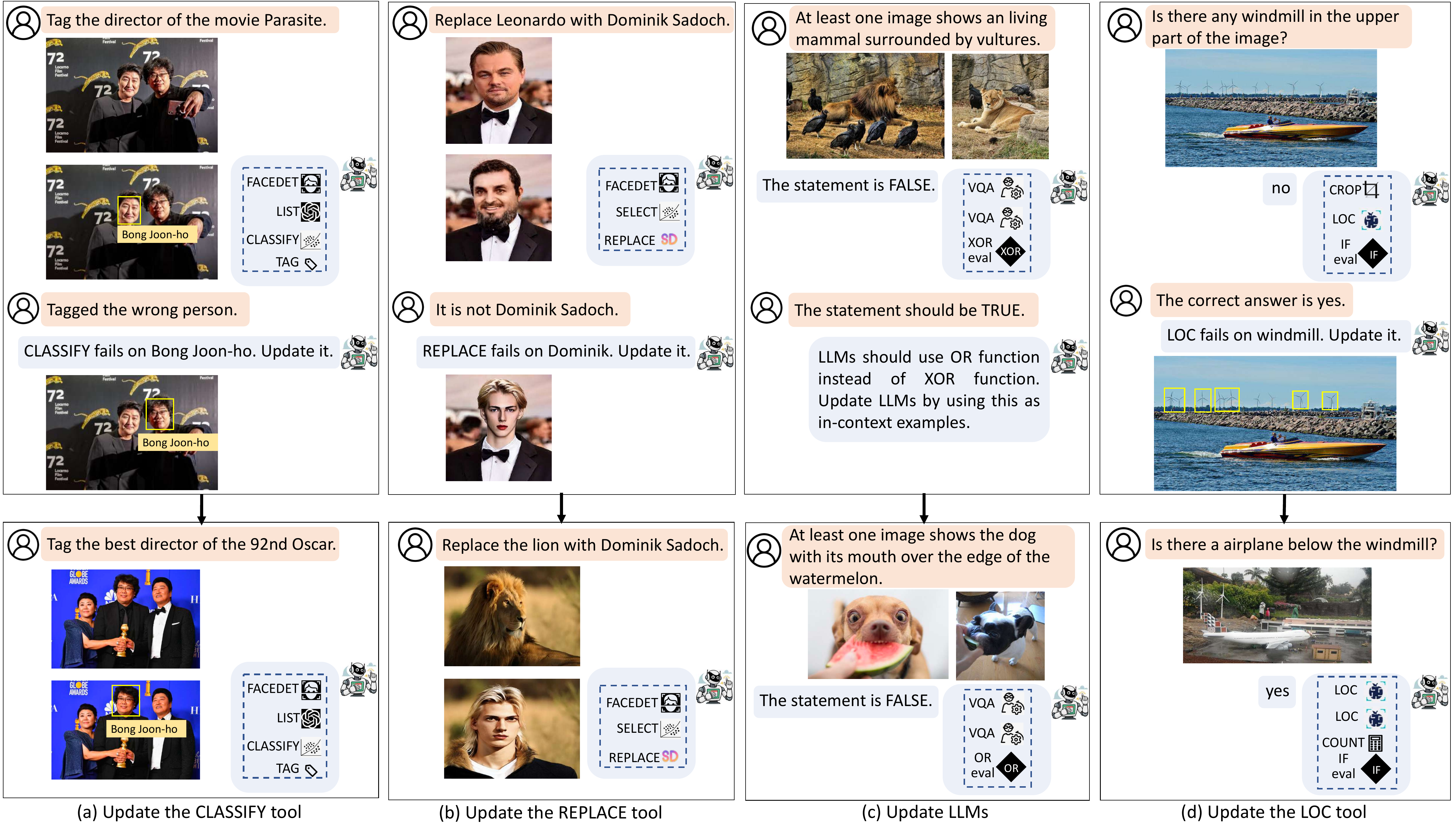}
    \vskip -0.08in
    \captionof{figure}{CLOVA is a general visual assistant that updates both LLMs and visual tools via inference, reflection, and learning in a closed-loop framework. During inference, CLOVA uses LLMs to integrate visual tools to accomplish given tasks. In reflection, CLOVA identifies tools that require updating based on human feedback. Finally, in learning, CLOVA collects data and updates the tools accordingly.
    }
    \label{fig:illustration_examples}
\end{center}%
}]
\renewcommand{\thefootnote}{\fnsymbol{footnote}} 
\footnotetext{\Letter~ Corresponding author: Qing Li (dylan.liqing@gmail.com).}

\begin{abstract}
\vskip -0.1in 
Utilizing large language models (LLMs) to compose off-the-shelf visual tools represents a promising avenue of research for developing robust visual assistants capable of addressing diverse visual tasks. However, these methods often overlook the potential for \textbf{continual learning}, typically by freezing the utilized tools, thus limiting their adaptation to environments requiring new knowledge. To tackle this challenge, we propose CLOVA, a \underline{C}losed-\underline{LO}op \underline{V}isual \underline{A}ssistant, which operates within a framework encompassing inference, reflection, and learning phases. During the inference phase, LLMs generate programs and execute corresponding tools to complete assigned tasks. In the reflection phase, a multimodal global-local reflection scheme analyzes human feedback to determine which tools require updating. Lastly, the learning phase employs three flexible approaches to automatically gather training data and introduces a novel prompt tuning scheme to update the tools, allowing CLOVA to efficiently acquire new knowledge. Experimental findings demonstrate that CLOVA surpasses existing tool-usage methods by 5\% in visual question answering and multiple-image reasoning, by 10\% in knowledge tagging, and by 20\% in image editing. These results underscore the significance of the continual learning capability in general visual assistants.
\end{abstract}

\section{Introduction}
\label{sec:intro}

The advancement of large language models (LLMs) \cite{touvron2023llama,abs-2303-08774} alongside various visual tools (e.g., neural networks and OpenCV functions) \cite{radford2021learning,rombach2022high,cheng2021per,minderer2022simple,li2022blip} offers feasible avenues for constructing general visual assistants. When confronted with a task accompanied by language instructions, a common approach involves harnessing LLMs to generate programs, which are then executed using readily available visual tools to solve the task as dictated by the generated program \cite{li2020competence,gupta2023visual,gao2023assistgpt,lu2023chameleon,surismenon2023vipergpt,shen2023hugginggpt}. For instance, when posed with the query "\textit{What is the person to the left of the umbrella doing?}", a viable solution entails LLMs sequentially performing the following steps: (1) utilizing a detection tool to locate the umbrella, (2) cropping the image region to the left of the umbrella, (3) using the detection tool to locate the person, and (4) finally querying a Visual Question Answering (VQA) tool with the question "\textit{What is the person doing?}". 
Being highly compositional, such tool-usage methods demonstrate impressive performance and appealing explainability in tackling complex reasoning tasks, including VQA~\cite{li2018vqa,li2018tell,gurari2018vizwiz}, mathematical reasoning~\cite{hong2021learning,hong2021smart}, and image editing~\cite{yang2023gpt4tools,yang2023mm,liu2023internchat,wu2023visual}.

However, the potential for continual learning has been largely overlooked in existing tool-usage methods. Most of them simply freeze the used tools, which limits their applicability to environments where new knowledge is required, as depicted in \cref{fig:illustration_examples}. For instance, a user might instruct a visual assistant to label the face of the movie director Bong Joon-ho in a photograph. However, if the face recognition tool employed by the assistant fails to recognize Bong Joon-ho, it may provide an incorrect response. In such scenarios, it is expected that the assistant can learn this missing information about Bong Joon-ho and generalize it to other photographs. Thus, it is imperative to endow visual assistants with the learning capability, enabling them to swiftly acquire new knowledge from failures.

In this paper, we propose CLOVA, a \underline{C}losed-\underline{LO}op \underline{V}isual \underline{A}ssistant that updates used tools via closed-loop learning \cite{li2020closed,li2024neural} to better adapt to new environments, as illustrated in Figure \ref{fig:illustration_examples}. CLOVA consists of three phases: inference, reflection, and learning. During inference, CLOVA employs Large Language Models (LLMs) to generate programs and execute corresponding tools to accomplish the task at hand. Subsequently, in the reflection phase, CLOVA utilizes human feedback to provide critiques, identifying tools that require updates. Finally, in the learning phase, CLOVA autonomously collects data and updates tools accordingly. Thus, CLOVA facilitates autonomous tool updating, thereby continually enhancing their ability to adapt to diverse environments.

To establish such a closed-loop learning framework, we must address three key challenges. Firstly, identifying tools that require updates is difficult due to the multi-step nature of generated programs and the diversity of errors within them. Secondly, automatically collecting training data is necessary as the knowledge to be learned is unpredictable. Thirdly, efficiently updating tools presents another obstacle, considering their scale and the quality of the collected data. Visual tools typically involve large neural networks, making them inefficient to update, and naive fine-tuning could result in unacceptable catastrophic forgetting~\cite{Luo2023AnES}. Moreover, the presence of noise within the collected data further complicates the training process.

We propose several techniques to tackle these challenges. First, we introduce a multimodal global-local reflection scheme, which resorts to LLMs to identify tools that need to be updated from both global and local aspects. For the second challenge, three data collection manners are employed, including inferring answers by LLMs, searching on the Internet, and searching from open-vocabulary datasets. Lastly, we develop a training-validation prompt tuning scheme for the tools, which includes instance-wise prompt tuning and a subsequent prompt validation stage, where learned prompts that fail to predict the validation data will be discarded. The learning phase also updates LLMs by storing correct examples and incorrect examples with critiques as in-context examples, which will be used in future inference. As a result, CLOVA efficiently updates tools in challenging environments with noisy data, while avoiding catastrophic forgetting.

We apply CLOVA to compositional VQA and multiple-image reasoning tasks, using the GQA~\cite{HudsonM19} and NLVRv2~\cite{SuhrZZZBA19} datasets. Additionally, we manually collect data for image editing and factual knowledge tagging tasks. CLOVA outperforms existing tool-usage methods by 5\% in compositional VQA and multiple-image reasoning tasks, by 10\% in knowledge tagging tasks, and by 20\% in image editing tasks, showing the significance of the learning capability for general visual assistants.

In summary, our contributions are three-fold:
\begin{itemize}[]
\item We build CLOVA, a visual assistant that updates its tools within a closed-loop learning framework for better adaptation to new environments.
\item We propose a multimodal global-local reflection scheme, capable of identifying tools in need of updates. \item We employ three flexible manners to automatically collect training data and introduce a novel training-validation prompt tuning scheme to update tools efficiently while avoiding catastrophic forgetting.
\end{itemize}

\section{Related Work}
\label{sec:relatedwork}


\subsection{General Visual Assistant}
Benefiting from the advancements of LLMs~\cite{touvron2023llama,abs-2303-08774} and visual tools~\cite{radford2021learning,Kirillov_2023_ICCV,minderer2022simple,rombach2022high}, visual assistants have achieved great progresses. Some methods concatenate and train LLMs with visual tools in an end-to-end manner, where representative work includes LLaVA~\cite{liu2023visual}, Otter~\cite{li2023otter}, MMICL~\cite{zhao2023mmicl}, Kosmos-2~\cite{peng2023kosmos}, and Flamingo~\cite{alayrac2022flamingo}, \emph{etc}. In addition, some work extends the idea of tool usage for AI assistants from natural language processing~\cite{qin2023tool,Paranjape2023ARTAM,Cai2023LargeLM,Qiao2023MakingLM,Schick2023ToolformerLM,hao2023toolkengpt} to computer vision. By providing in-context examples, VISPROG~\cite{gupta2023visual} and ViperGPT~\cite{surismenon2023vipergpt} generate programs to use visual tools. Following this idea, some work improves performance by collecting instruction-following data~\cite{yang2023gpt4tools,liu2023llavaplus,patil2023gorilla}, adding more tools~\cite{shen2023hugginggpt,liang2023taskmatrix}, and designing more dedicated tool-usage procedures~\cite{xu2023ullava,yang2023mm,2023controlllm,lu2023chameleon,wu2023visual,liu2023internchat,hu2023tree}. The most related work to CLOVA is AssistGPT~\cite{gao2023assistgpt} and OpenAGI~\cite{ge2023openagi}. The two methods update LLMs after development through in-context learning and reinforcement learning, respectively. Different from them, CLOVA can update both LLMs and visual tools via its reflection and learning phases. This allows CLOVA to better adapt to new environments. In addition, the closed-loop framework enables us to set a separate training stage for tool-usage methods, going beyond zero-shot or few-shot visual assistants. Comparisons between CLOVA and some representative tool-usage methods are shown in~\cref{tab:related_work}.

\begin{table}
  \centering
  \resizebox{1\columnwidth}{!}{
  \begin{tabular}{c |c c c c}
    \toprule
    {Method} & Visual Tool & Reflection & Update LLMs & Update VTs\\
    \hline
    ART~\cite{Paranjape2023ARTAM} & \xmark  & \xmark & Prompt & - \\
    TRICE~\cite{Qiao2023MakingLM} & \xmark  & Global & Instruction + RL & - \\
    ToolkenGPT~\cite{hao2023toolkengpt} & \xmark  & - & \xmark & -  \\
    Toolformer~\cite{Schick2023ToolformerLM} & \xmark & - & Fine-tune & -  \\
    \hline
     VISPROG~\cite{gupta2023visual} &\cmark  & \xmark  & \xmark & \xmark \\
     Visual ChatGPT~\cite{wu2023visual} &\cmark & \xmark &\xmark & \xmark\\
     HuggingGPT~\cite{shen2023hugginggpt}  &\cmark & \xmark & \xmark & \xmark \\
     ViperGPT~\cite{surismenon2023vipergpt} &\cmark & \xmark & \xmark &\xmark\\
     GPT4TOOLs~\cite{yang2023gpt4tools} &\cmark & \xmark & Instruction  & \xmark\\
     OpenAGI~\cite{ge2023openagi} &\cmark & \xmark & RL & \xmark \\
     AssistGPT~\cite{gao2023assistgpt} &\cmark & Global & Prompt & \xmark\\
     \midrule
     \textbf{CLOVA (Ours)}&\cmark & Global+Local & Prompt & Prompt\\
    \bottomrule
  \end{tabular}
  }
  \caption{Comparisons with representative tool-usage methods, where VTs means visual tools.
  }
  \vskip -0.2in
  \label{tab:related_work}
\end{table}

\subsection{Reflection of LLMs}

Reflection has become a remedy in case LLMs cannot generate good responses in a single attempt~\cite{pan2023automatically,yang2022re3,park2023generative,chen2023teaching}. Reflection methods send outputs back to LLMs to obtain critiques and further improve the outputs. These critiques take the form of scores or natural language~\cite{madaan2023self,shinn2023reflexion,nair2023dera,abs-2310-03051}. To generate better critiques, some methods employ instruction tuning~\cite{yan2023learning,saunders2022self} or reinforcement learning~\cite{Qiao2023MakingLM,ConstitutionalAI}. Recently, Huang \emph{et al.}~\cite{huang2023large} revealed that LLMs struggle to provide accurate critiques for complex tasks. One way to address this issue is incorporating external information such as human-desired results into LLMs~\cite{yang2022re3,xu2023search,abs-2310-11511}. Unlike existing methods that rely solely on feedback in the language modality, our method generates reflection using all multimodal intermediate results. In addition, our method incorporates both global and local aspects for reflection, instead of only the global aspect. These bring more effective critiques for compositional tasks.

\subsection{Prompt-based Learning}

Prompt-based learning is an efficient technique to update neural networks, achieving impressive performance in both NLP~\cite{shin2020autoprompt,jiang2020can} and computer vision~\cite{zhu2023visual,Shao_2023_CVPR,zhu2023segprompt,Huang_2023_CVPR}. Prompt engineering and prompt tuning are two kinds of commonly used methods. Prompt engineering develops interpretable tokens (\emph{e.g.}, texts and image regions) to guide model prediction, which are usually obtained by manually designing~\cite{tang2023cotdet,radford2021learning},  retrieval~\cite{zhang2023VisualPromptRetrieval}, and model generation~\cite{hu2022scaling,pratt2023does}. Prompt tuning learns vectors as prompts via gradient-based optimization.  VPT~\cite{jia2022visual} and CoOp~\cite{zhou2022learning} learn prompts for vision encoders and text encoders, respectively. To handle diverse data, CoCoOp~\cite{zhou2022conditional} learns to generate prompts for unknown classes, ProDA~\cite{lu2022prompt} builds a Gaussian distribution for prompts, and MaPLe~\cite{khattak2023maple} learns both text and visual prompts. In addition, some methods employ prompts for continual learning, which learn prompts for different classes and produce adaptive prompts during inference. Representative methods include PIVOT~\cite{villa2023pivot}, DualPrompt~\cite{wang2022dualprompt}, and L2P~\cite{wang2022learning}. Different from existing methods, our training-validation prompt tuning scheme discards harmful prompts, leading to more stable learning processes when the quality of training data is subpar.
\begin{figure}
    \centering
    \includegraphics[width=0.48\textwidth]{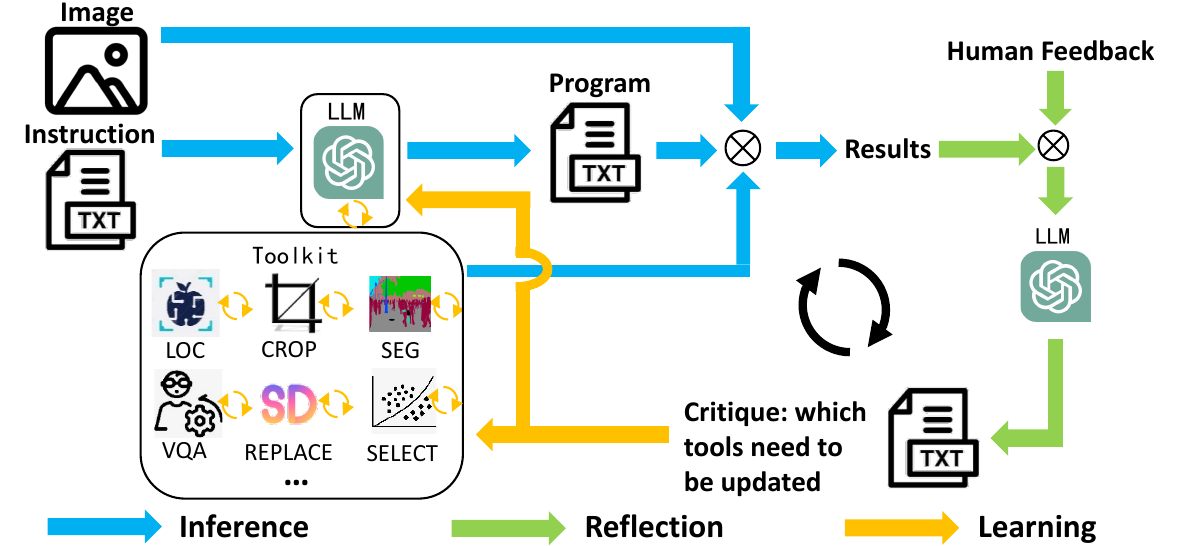}
    \caption{Framework of CLOVA.}
      \vskip -0.2in
\label{fig:framework_colva}
\end{figure}

\section{Method}
\label{sec:method}

\subsection{Overview}

As shown in~\cref{fig:framework_colva}, CLOVA has three phases: inference, reflection, and learning. In the inference phase, CLOVA uses LLMs to generate programs and executes corresponding tools to solve the task. The reflection phase introduces a multimodal global-local reflection scheme that uses LLMs to generate critiques, identifying which tool needs to be updated. During learning, we employ three manners to collect training data and use a training-validation prompt tuning scheme to update the tools.

\begin{table*}
\small
  \centering
  \resizebox{2\columnwidth}{!}{
  \begin{tabular}{c|c|c|c}
    \hline
    \bfseries Tool Type & \bfseries Tool Name & \bfseries Tool Description & \bfseries Data Collection\\
    \hline
    \multirow{6}{*}{\makecell{Tools \\ to be \\ updated}} & LOC & Use the OWL-ViT model~\cite{minderer2022simple} for object localization & Open-vocabulary datasest\\
    & VQA & Use the BLIP model~\cite{li2022blip} for VQA & LLM inference \\
    & SEG & Use the maskformer model~\cite{cheng2021per} for panoptic segmentation & Open-vocabulary datasest\\
    & SELECT & Use the CLIP model~\cite{radford2021learning} to select the most relevant object, given a text description & Internet\\
    & CLASSIFY & Use the CLIP model~\cite{radford2021learning} to classify given images & Internet \\
    & REPLACE & Use the stable diffusion inpainting model~\cite{rombach2022high} to replace one object with another desirable object & Internet\\
    \hline
    \multirow{9}{*}{\makecell{Tools \\ not to be \\ updated}} 
    & FACEDET & Use the DSFD model~\cite{li2019dsfd} for face detection & N/A \\
    & LIST & Use the text-davinci-002 model of OpenAI for knowledge retrieval & N/A \\
    & EVAL & Use the Python function eval() to process string expressions for answers & N/A\\
    & RESULT & Use the Python function dict() to output the intermediate and final results & N/A\\
    & COUNT & Use Python function len() to count the number of input bounding boxes or masks & N/A\\
    & CROP & Use Python function PIL.crop() to crop images & N/A\\
    & COLORPOP & Use Python function PIL.convert() to keep desirable objects in color and other regions gray & N/A\\
    & BGBLUR & Use Python function PIL.GaussianBlur() to blur the background & N/A \\
    & EMOJI & Use emojis in the Python packages AngLy(pypi) to hide someone’s face & N/A\\
    \hline
  \end{tabular}
  }
  \vskip -0.1in
  \caption{Used tools in CLOVA, categorized based on whether the tool is updated in our method.  Details of tool updates are in \cref{sec:tool_update}}
    \vskip -0.2in
  \label{tab:toolexample}
\end{table*}

\subsection{Inference}
\begin{figure}
    \centering
    \includegraphics[width=0.45\textwidth]{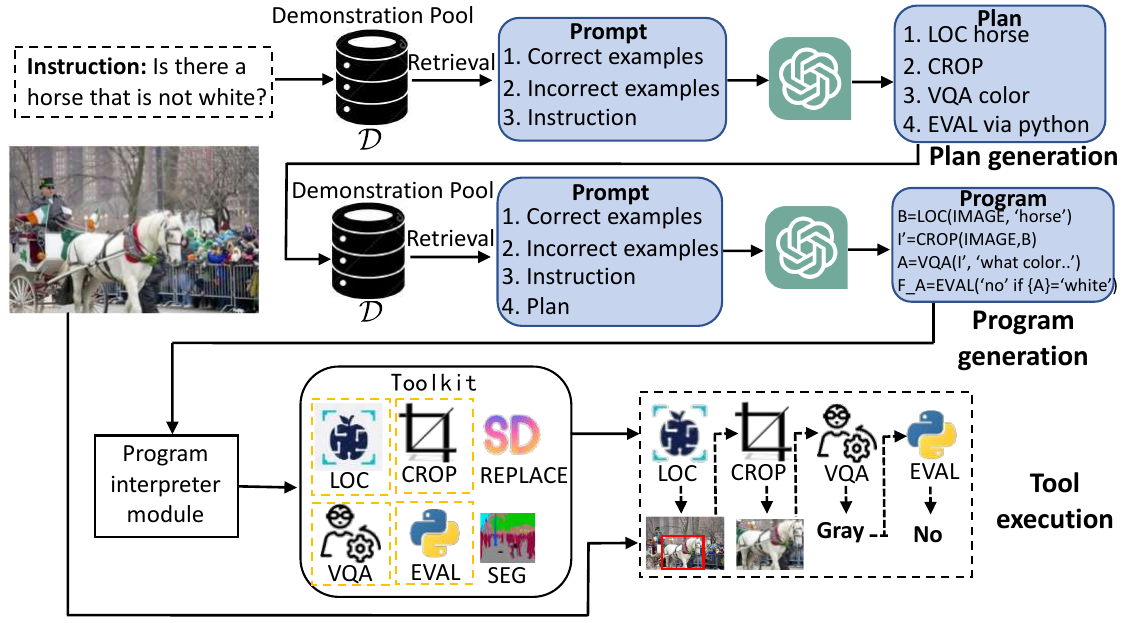}
    \caption{Illustration of the inference phase in CLOVA.}
      \vskip -0.2in
\label{fig:framework_inference}
\end{figure}

Our inference phase is based on VISPROG~\cite{gupta2023visual}, while the difference is that CLOVA first uses LLMs to generate plans and then generates programs based on the plans, instead of directly generating programs. Plans can be seen as intermediate reasoning chains that benefit the inference and reflection phases. Given a task, CLOVA selects in-context examples from a demonstration pool $\mathcal{D}$ (the construction of $\mathcal{D}$ will be detailed in \cref{sec:llm_learning}), including correct examples and incorrect examples with error critiques. These examples are used to create prompts that are then sent to LLMs for plan and program generation. Finally, the program is parsed to execute visual tools (see~\cref{fig:framework_inference}).

\noindent
\textbf{Plan generation.}
The demonstration pool $\mathcal{D}$ is composed by $\mathcal{D}=\{\mathcal{D}_{p,s},\mathcal{D}_{p,f},\mathcal{D}_{c,s},\mathcal{D}_{c,f} \}$, where $\mathcal{D}_{p,s}$ and $\mathcal{D}_{p,f}$ contain correct and incorrect examples for plan generation respectively, and $\mathcal{D}_{c,s}$ and $\mathcal{D}_{c,f}$ contain correct and incorrect examples for program generation respectively. Given a task, we use the BERT model~\cite{kenton2019bert} to extract features of the given instruction and examples stored in $\mathcal{D}_{p,s}$ and $\mathcal{D}_{p,f}$. Then, we combine similar examples in $\mathcal{D}_{p,s}$ and $\mathcal{D}_{p,f}$ with the instruction to create a prompt. Finally, we send the prompt to LLMs to generate the plan in a one-go manner. 

\noindent
\textbf{Program generation.}
Similar to plan generation, we use LLMs to generate programs in a one-go manner.
We select correct and incorrect examples of programs from $\mathcal{D}_{c,s}$ and $\mathcal{D}_{c,f}$. 
We combine these examples with the plan as a prompt and then send the prompt to LLMs for program generation.


\begin{figure}
    \centering
    \includegraphics[width=0.45\textwidth]{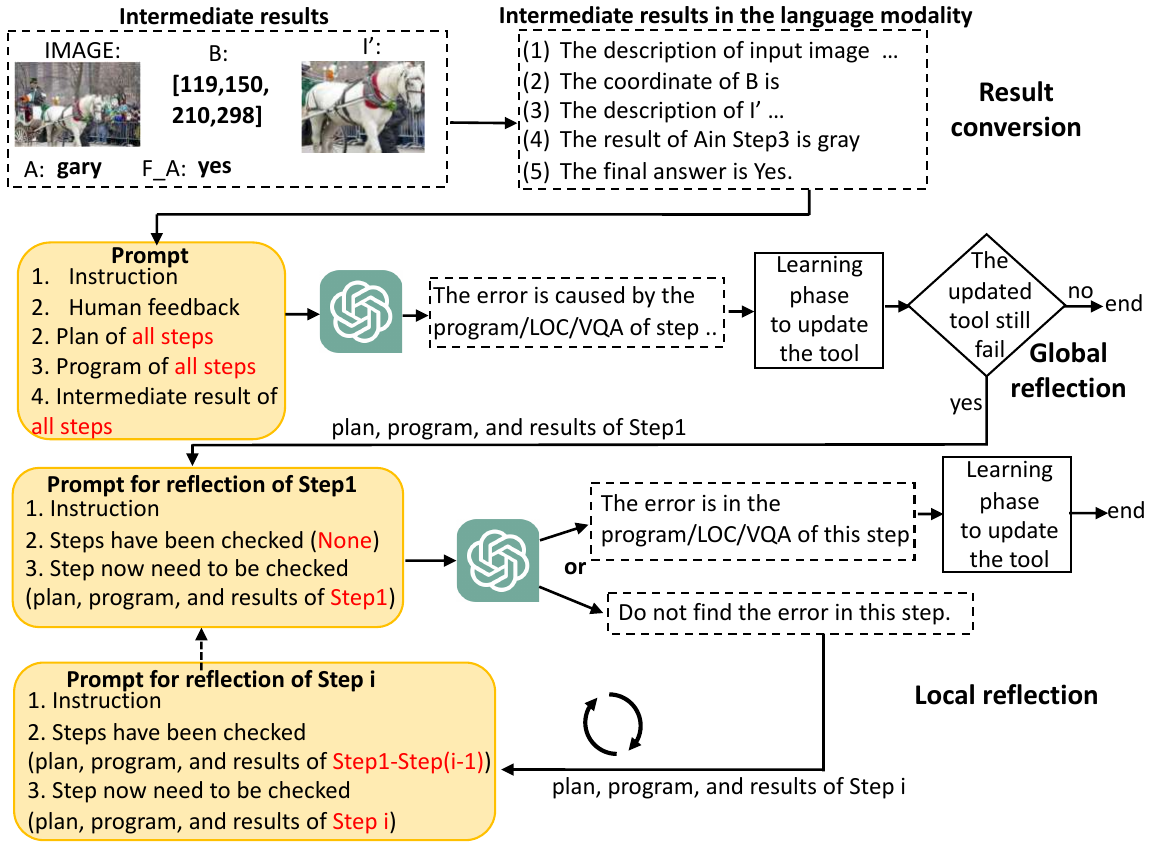}
    \caption{Illustration of the reflection phase in CLOVA. 
}
  \vskip -0.2in

\label{fig:framework_reflection}
\end{figure}

\noindent
\textbf{Tool execution.}
We utilize the interpreter module in~\cite{gupta2023visual} to parse the program, extracting tool names, inputs, and outputs of each step. CLOVR activates tools from a toolkit $\mathcal{T}$ that contains $15$ tools, including neural networks and Python functions, as shown in~\cref{tab:toolexample}.

\subsection{Reflection}

In the inference phase, if a task is not solved correctly, the multimodal global-local reflection scheme uses LLMs to generate critiques, identifying which tool needs to be updated, as shown in~\cref{fig:framework_reflection}.

\noindent
\textbf{Result conversion.}
Since LLMs often struggle to identify errors by themselves~\cite{huang2023large,stechly2023gpt4,valmeekam2023large}, we provide the human feedback, our wrong results, and intermediate results of each step for LLMs to better identify the error source. This requires us to convert visual results into textual form. For this purpose, we use the BLIP~\cite{li2022blip} model to convert images into languages by captioning.

\noindent
\textbf{Global reflection.}
CLOVA first uses global reflection to generate critiques in a one-go manner. The prompts are composed of task inputs, feedback on the task (\emph{e.g.}, desirable results in VQA tasks, or human comments in image editing tasks), generated plans and programs, and intermediate results at each step. We send the prompts to LLMs to generate critiques that are used to update tools in the learning phase.

\noindent
\textbf{Local reflection.}
If CLOVA still fails after the tools are updated via global reflection and the learning phase--meaning the actual tools that lead to the faulty response are still to be found, we resort to local reflection to analyze each step of the program. Prompts are composed of the task inputs, feedback on the task, the steps that have been checked, and the current step that needs to be checked. Each step includes plans, programs, and intermediate results. We send the prompts to LLMs to infer whether this step has errors and the reasons. Local reflection continues until an error location and reasons are identified for a step.

\subsection{Learning}\label{sec:tool_update}

\subsubsection{Updating tools with prompt tuning}

\begin{figure}
    \centering
    \includegraphics[width=0.47\textwidth]{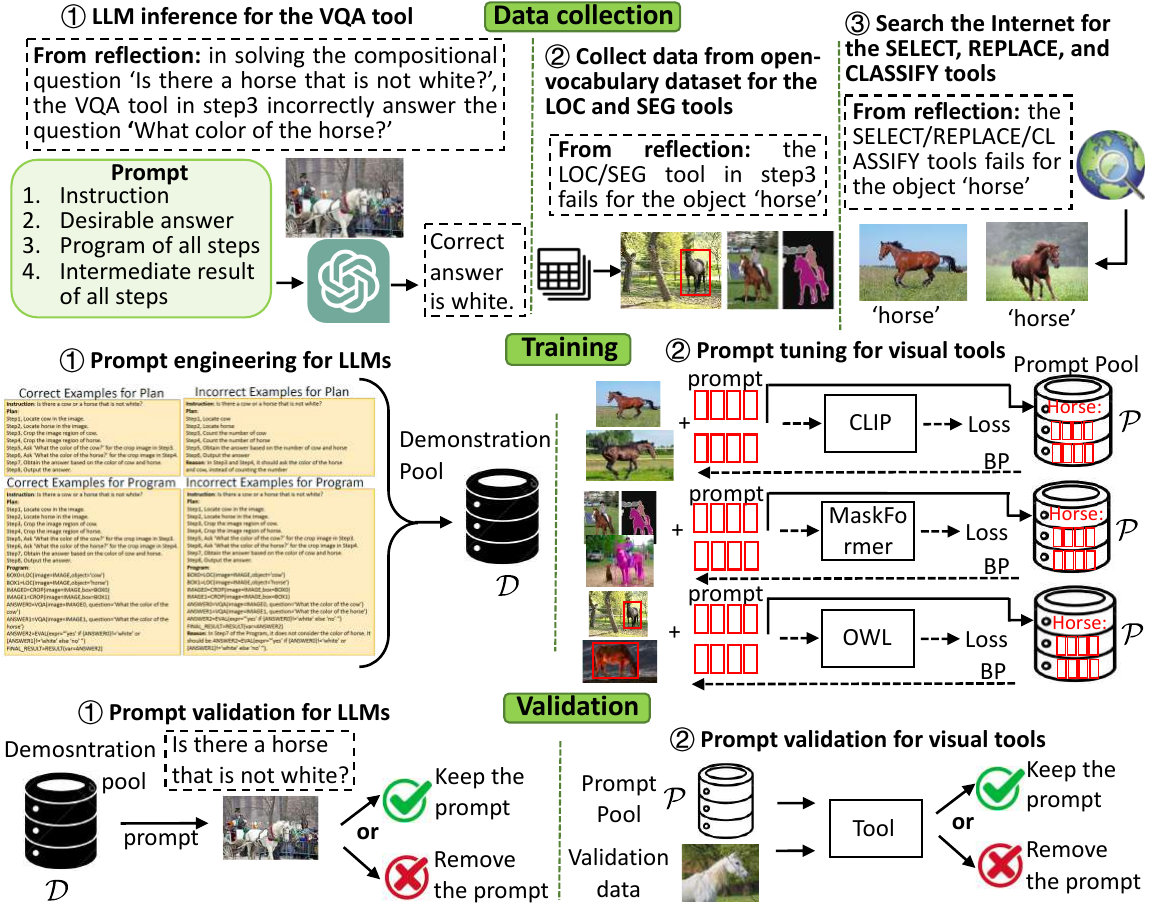}
    \caption{Illustration of the learning phase in CLOVA. 
    }
      \vskip -0.2in
\label{fig:framework_learning}
\end{figure}

After identifying tools that need to be updated from the reflection phase, CLOVA then moves to the learning phase to collect training data and goes through training-validation prompt tuning to update the tools, as shown in~\cref{fig:framework_learning}.

\noindent
\textbf{Data collection.}
Since the tools that need to be updated can be rather different (a full list can be found in \cref{tab:toolexample}), we explore three manners to collect data online. 
(1) We use LLMs to generate training data for the VQA tool. If reflection concludes that the VQA tool makes errors, we combine the desirable response of the whole task and intermediate results of each step to prompt LLMs into inferring the correct output of the VQA tool. The question and the inferred output are then used to update the VQA tool.
(2) We gather training data from open-vocabulary visual object grounding datasets (\emph{e.g.}, LVIS~\cite{gupta2019lvis}) for the LOC and SEG tools. For example, if the reflection phase indicates that LOC does not work well for the visual concept ``\emph{horse}'', CLOVA will select images and bounding boxes of horses from LVIS to update LOC. 
(3) We collect data by searching on the Internet for the SELECT, CLASSIFY, and REPLACE tools. For instance, 
If CLASSIFY is marked as unable to recognize ``\emph{horse}'' during the reflection phase, CLOVA will search the Internet for images of horses to update CLASSIFY.

\noindent
\textbf{Prompt tuning and validation.}
Given the collected data, we invoke training-validation prompt tuning to update tools. Note that, instead of learning a single prompt for all collected data, we choose to learn a prompt for each training instance collected. Each learned prompt will then be validated by running the tool with it on validation data (held out from collected data except for the VQA tool, where VQA will be validated on the original visual question it failed on) and seeing if the tool can produce desirable responses (\textit{e.g.} correctly localizing a horse for the LOC tool). As a result, we discard prompts that do not lead to the desirable responses, possibly due to the faulty training instances they were trained on, alleviating the issue of noisy collected data. Finally, we build a prompt pool $\mathcal{P}$ for each tool. Take the LOC tool as an example. After training and validation, CLOVA stores the visual concept (\textit{e.g.}, ``horse'' in the task of localizing horses) with its learned prompts and visual features of all collected instances in $\mathcal{P}$, formulated as $\mathcal{P} = \left\{v_j: \big[  [f_{j1}, \cdots, f_{jn}], [p_{j1},\cdots, p_{jn}] \big] \right\} _{j=1}^{m}$, where $v_j$ is the name of the $j$-the concept (\emph{e.g.}, ``horse''), $m$ concepts are stored totally. For the concept $v_j$, $f_{ji}$ and $p_{ji}$ are the feature and learned prompt using the $i$-th instance, respectively, and $n$ instances are learned for $v_j$.

In summary, a tool is formulated as $T_{\theta, \mathcal{P}}$, where $\theta$ is the parameter of neural networks. The forward process of a sample $x$ is $T_{\theta, \mathcal{P}}(x)=\theta([x,\mathcal{P}])$, where we concatenate $x$ with a retrieved prompt from $\mathcal{P}$ as the input for the tool. We update the tool $T_{\theta, \mathcal{P}}$ by tuning $\mathcal{P}$ while $\theta$ being fixed, 
\vskip -0.12in
\begin{equation*}
\begin{aligned}
\min_{\mathcal{P}} \mathbb{E}_{(x,y)} \mathcal{L}\big( T_{\theta, \mathcal{P}} (x),y  \big),
\end{aligned}
\end{equation*}
\vskip -0.1in
\noindent where $(x,y)$ is collected data, $\mathcal{L}$ is the loss function of $T_{\theta, \mathcal{P}}$.

\noindent
\textbf{Prompt ensemble.}
During inference, we use prompt ensemble to retrieve and utilize prompts from the learned prompt pool $\mathcal{P}$. Specifically, given a generated program, we first identify the visual concept for each involved tool in the program. For example, given an image editing task, ``\emph{Replace the dog with a cat}'' where the SEG, SELECT, and REPLACE tools will be used, ``\emph{dog}'' is the visual concept for the SEG and SELECT tools, and ``\emph{cat}'' is the visual concept for the REPLACE tool. Then, in each step of tool usage with an input image $x$, if the visual concept is not in $\mathcal{P}$ of the corresponding tool, the prompt $p'$ for $x$ will be set as a zero-vector, \textit{i.e.} the concept has not been learned for the tool so we just use the original tool (using zero-vector as a prompt); if the visual concept can be found in $\mathcal{P}$, \textit{i.e.} the tool was updated with the visual concept before, we aggregate the prompts corresponding to this concept based on the similarity between features stored with these prompts and the feature extracted from the current input $x$: $p'$ is computed by $p' = \frac{\sum_{i=1}^{n} w_i \cdot p_{ji}}{\sum_{i=1}^{n} w_i}$, where we compute the cosine similarity between feature $f_x$ of $x$ and features $f_{ji}$ in $\mathcal{P}$ as the weight $w_i$.

\vskip -1 in
\subsubsection{Updating LLMs with demonstrations}
\label{sec:llm_learning}

Besides the visual tools, CLOVA can also update its LLMs. As we mentioned above, CLOVA utilizes a demonstration pool $\mathcal{D}$ to provide relevant examples for the LLMs. After working on new data, the plan, program, and reflection will be stored in $\mathcal{D}$ as correct or incorrect examples, based on whether the data is correctly proceeded. We also have a validation process that uses the original instruction as validation data to evaluate stored in-context examples. As the size of $\mathcal{D}$ grows, LLMs use more examples and therefore strengthen reasoning skills.

\section{Experiments}
\label{sec:experiments}


\subsection{Setting}

Following VISPROG~\cite{gupta2023visual}, we evaluate our method on four tasks: {compositional VQA}, {multiple-image reasoning}, {language-guided image editing}, and {factual knowledge tagging}, which requires visual perception, compositional reasoning, and image generation and manipulation abilities.

To comprehensively evaluate the learning capability of CLOVA, we set a separate training stage before deployment, which iteratively learns new knowledge via inference, reflection, and updating phases. In the test stage (\emph{i.e.}, after development), we do not update LLMs and visual tools, and evaluate the performance only via the inference phase. 

In the compositional VQA task, the GQA dataset~\cite{HudsonM19} is used. We randomly select $500$ samples from its train split as the training data, and $500$ samples from its test-dev split as the test data. We report the top-1 accuracy. In the multiple-image reasoning task, we use the NLVRv2 dataset~\cite{SuhrZZZBA19} that provides two images and a statement. We need to judge whether the statement is true or false. Similarly, we randomly select $500$ samples from its train split as our training data, and $500$ samples from its dev split as our test data.

Similar to VISPROG, we manually collect data for the language-guided image editing and factual knowledge tagging tasks. To better evaluate the learning capability, we collect fine-grained visual concepts that visual tools may not have learned, such as ``\emph{Replace the lion in the image with pine grosbeak}'', where pine grosbeak is a fine-grained bird of Passeriformes. In the image editing task, we collect $129$ images with $193$ instructions, where $27$ images with $78$ instructions are used for training, and the rest are test data. We manually check whether edited images are semantically correct. The factual knowledge tagging task needs to identify persons or objects with bounding boxes in images. We collect $86$ images with $177$ instructions for this task, where $10$ images with $88$ instructions are used for training and the rest are used as the test data. We report the F1 score for this task.

In plan and program generation, prompts contain $4$ correct examples and $4$ incorrect examples. The demonstration pool $\mathcal{D}$ is initialized having about $20$ in-context examples. 

\begin{figure*}
    \centering
    \includegraphics[width=0.95\textwidth]{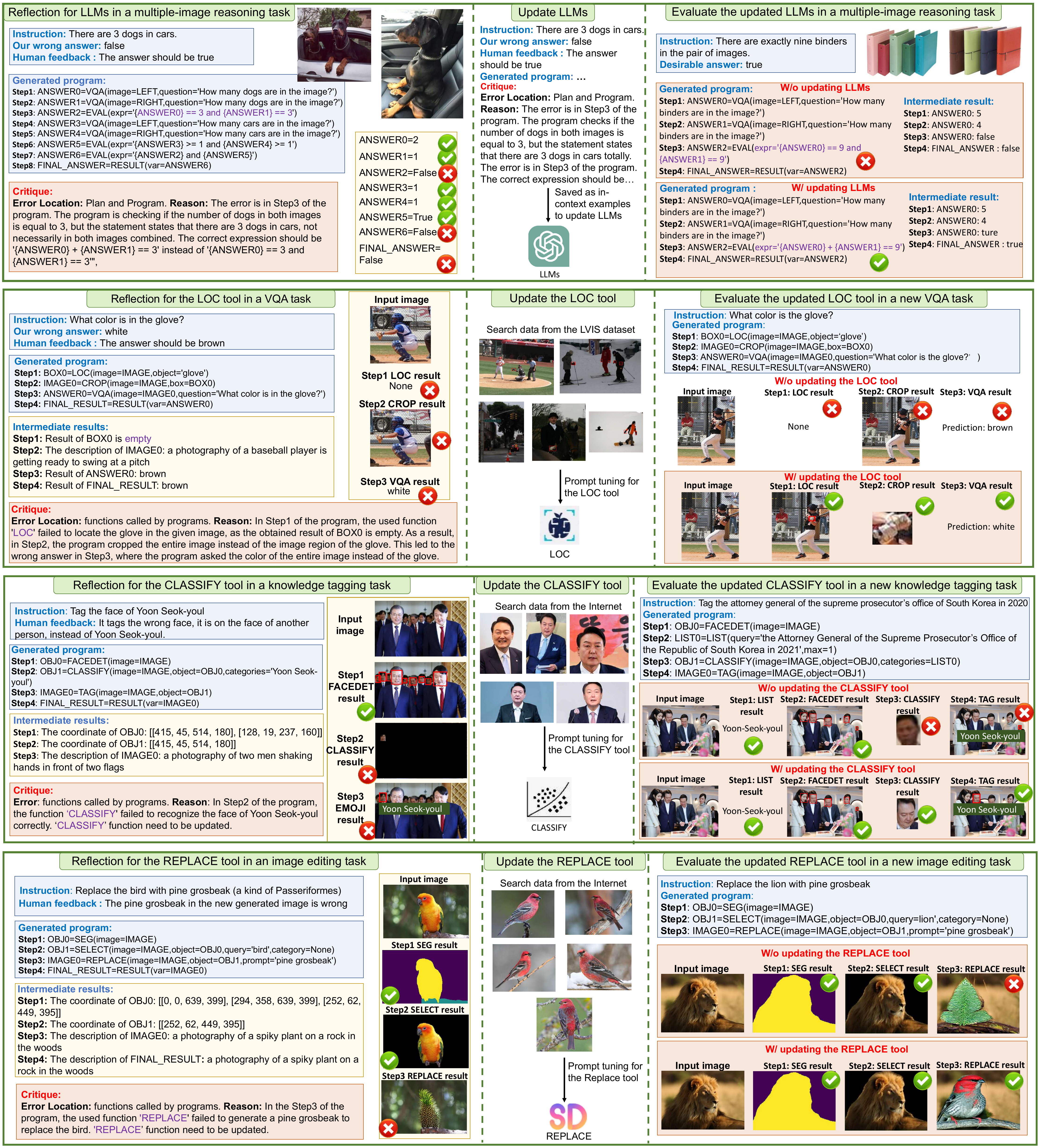}
  \vskip -0.1in
      \caption{Case study of CLOVA on four example tasks.}
\label{fig:learning_all}
  \vskip -0.2in
\end{figure*}

\subsection{Main Results}

\begin{table}
  \centering
  \Large
  \resizebox{1\columnwidth}{!}{
  \begin{tabular}{c | c | c c c c}
    \hline
      & Method  & GQA & NLVRv2 & Editing &  Tagging \\
    \hline
    \multirow{2}{*}{E2E} & Otter~\cite{li2023otter} & $48.2$ & $48.2$ & - & - \\
    & MMICL~\cite{zhao2023mmicl} & $64.4$ & $62.2$ & - & - \\
    \hline
    \multirow{7}{*}{Tool} & GPT4TOOLs~\cite{wu2023visual} & $41.2$ &  $45.4$ & $17.8$ & \\
    & Visual ChatGPT~\cite{wu2023visual} & $43.2$ &  $51.6$ & $21.7$ & - \\
    & InternGPT~\cite{liu2023internchat} & $44.8$ & $39.4$ & - & -\\
    & HuggingGPT~\cite{shen2023hugginggpt} & $46.0$ & $44.0$ & - & -\\
    & ViperGPT~\cite{surismenon2023vipergpt} & $47.2$ & - & - & -\\
    & VISPROG~\cite{gupta2023visual} & $49.8$ & $60.8$ & $40.2$\ & $0.393$\\
    \cline{2-6}
    & CLOVA (Ours) & $\mathbf{54.6}$ & $\mathbf{65.6}$ & $\mathbf{65.4}$ & $\mathbf{0.502}$\\
    \hline
  \end{tabular}
  }
  \vskip -0.1in
  \caption{Comparisons in the four tasks. We report accuracies on GQA, NLVRv2, and image editing tasks, and F1 score on the knowledge tagging task. `E2E' means end-to-end methods.}
    \vskip -0.2in
  \label{tab:mainresultsgqa}
\end{table}

We compare CLOVA with tool-usage methods: VISPROG~\cite{gupta2023visual}, GPT4TOOLs~\cite{yang2023gpt4tools}, Visual ChatGPT~\cite{wu2023visual}, InternGPT~\cite{liu2023internchat}, HuggingGPT~\cite{shen2023hugginggpt}, and ViperGPT~\cite{surismenon2023vipergpt}. We use their official codes, where all methods use the GPT-3.5 model. In addition, we also compare CLOVA with two advanced end-to-end models: Otter~\cite{li2023otter} and MMICL~\cite{zhao2023mmicl}, which do well in GQA and NLVRv2 datasets. Results on the four tasks are shown in~\cref{tab:mainresultsgqa}. We observe that CLOVA achieves the best performance among tool-usage methods. CLOVA has at least $4.8 \%$ improvements on the GQA dataset and $4.9 \%$ improvements on the NLVRv2 dataset. The reason is that CLOVA learns how to generate programs for the two tasks and update the VQA and LOC tools for better image perception. CLOVA performs competitively and even outperforms Otter and MMICL.

\begin{figure*}
    \centering
    \includegraphics[width=1\textwidth]{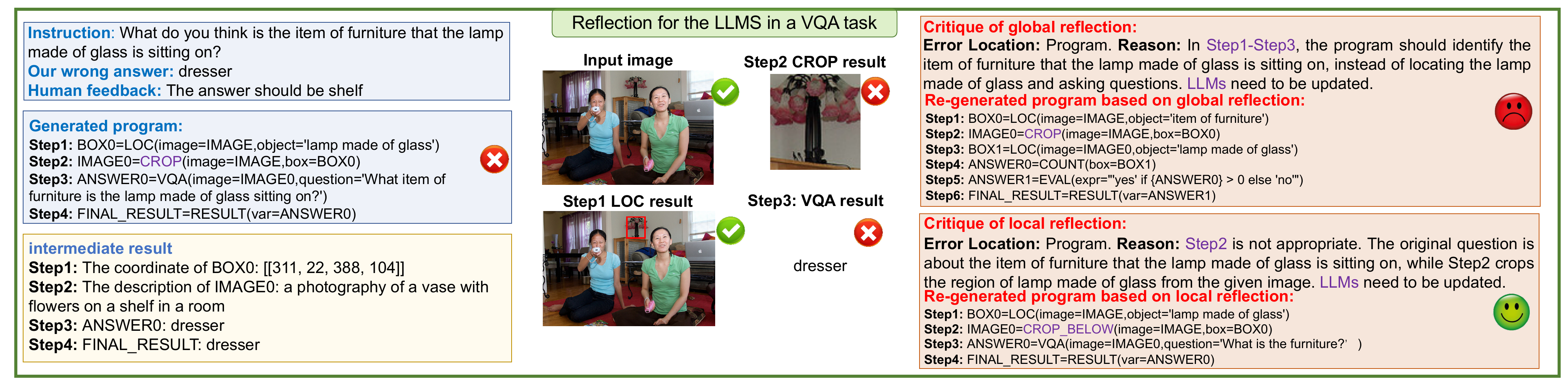}
  \vskip -0.1in
      \caption{Visualization of the global reflection and local reflection in a VQA task.}
\label{fig:reflection_all}
  \vskip -0.2in
\end{figure*}

On image editing and knowledge tagging tasks, we do not compare our method with InterGPT, HuggingGPT, and ViperGPT, since they either need object masks or cannot accurately locate objects. In addition, most methods cannot finish the tagging task. Thus, we compare our method with VISPROG and Visual ChatGPT. As we collect find-grained data, it is challenging for off-the-shelf classification, segmentation, and image generation tools. Since GPT4TOOLs and Visual ChatGPT cannot use OpenCV functions and do not have the learning capability, they get bad performance on the image editing task. VISPROG can use OpenCV functions, but it cannot learn new knowledge. Its main fault is the inability to recognize or generate fine-grained concepts. Compared with them, the learning capability of CLOVA brings more than $20 \%$ and $10 \%$ improvements to image editing and knowledge tagging tasks, respectively.

\subsection{Qualitative Results}

In~\cref{fig:learning_all}, we visualize four cases to illustrate the reflection and learning capability in CLOVA. It identifies tools that need to be updated, no matter LLMs or visual tools. Prompt engineering guides LLMs to generate correct programs for similar instructions. Visual tools learn new concepts via our data collection and prompt turning schemes. In~\cref{fig:reflection_all}, we visualize an example of global and local reflection. When the instruction is complex, the global reflection does not accurately identify which step has the error. Using global reflection as in-context examples still cannot generate correct programs. In contrast, local reflection successfully identifies the error step, and using local reflection generates correct programs, showing the effectiveness of local reflection.

\begin{table}
  \centering
  \resizebox{1\columnwidth}{!}{
  \begin{tabular}{c | c c c}
    \hline
    & Method & GQA & NLVRv2 \\
    \hline
    \multirow{5}{*}{Reflection}  
    & w/o local reflection & $52.0$ & $65.2$\\
    & w/o global reflection & $53.6$ & $64.2$ \\
    & w/o intermediate results & $48.8$ & $61.2$\\
    & w/o plan & $50.0$ & $62.6$\\
    & Ours & $\mathbf{54.6}$ & $\mathbf{65.6}$\\
    \hline
    \multirow{4}{*}{\makecell{Prompt Engineering  \\ for LLMs}} & w/o incorrect cases &  $46.1$  &  $61.4$ \\
    & w/o correct cases & $48.2$ & $63.2$ \\
    & w/o validation & $44.2$  & $61.0$ \\
    & Ours & $\mathbf{54.6}$ & $\mathbf{65.6}$\\
    \hline
    \multirow{2}{*}{\makecell{Prompt Tuning \\ for visual tools}} 
    & w/o validation & $42.8$ & $62.8$\\
    & Ours & $\mathbf{54.6}$ & $\mathbf{65.6}$\\
    \hline
  \end{tabular}
  }
  \vskip -0.1in
  \caption{Ablation on the GQA and NLVRv2 dataset.}
  \label{tab:ablation}
\end{table}

\subsection{Ablation Studies}
We conduct ablation studies on the reflection and learning phases, using the GQA and NLVRv2 datasets. For reflection, we evaluate only using global reflection, only using local reflection, not using multimodal intermediate results, and not generating plans. We separately evaluate learning schemes for LLMs and visual tools. We evaluate only storing correct examples or incorrect examples for updating LLMs. We also evaluate removing the validation process in prompt engineering and prompt tuning processes. Results are shown in~\cref{tab:ablation}. We find that these components are necessary for CLOVA to achieve better performance.

We evaluate CLOVA using different LLMs: LLaMA2-7B, GPT-3.5-turbo, and GPT-4. Results on GQA and NLVRv2 are shown in~\cref{tab:llmsgqa}. We find that CLOVA leads to improvements in both strong LLMs (GPT-4) and weaker LLMs (GPT-3.5 and LLaMA2-7). We observe that CLOVA even achieves higher improvements on open-source LLMs (\emph{i.e.}, LLaMA2-7B), $21 \%$ on the GQA dataset and $13.8 \%$ on the NLVRv2 dataset, bringing the significance of studying the learning capability of visual assistants. We further conduct experiments to evaluate CLOVA on two open-source LLMs: LLaMA2-7B and Mistral-7B. Results are shown in~\cref{table:opensource}. We observe that CLOVA achieves significant improvements again.

\begin{table}
  \centering
  \resizebox{1\columnwidth}{!}{
  \begin{tabular}{c | c |c c c}
    \hline
    Dataset  & Method & LLaMA2-7B & GPT-3.5 & GPT-4 \\
    \hline
    \multirow{3}{*}{GQA} & Baseline & $39.2$ & $46.4$ & $52.6$\\
   &  + Update LLMs & $56.8$ & $51.6$ & $56.6$\\
   & + Update visual tools & $60.2$ & $54.6$ & $60.4$ \\      
    \hline
    \multirow{3}{*}{NLVRv2}  & Baseline & $50.0$ & $60.2$ & $64.8$\\
   & + Update LLMs & $59.2$ & $63.6$ & $68.8$\\
   & + Update visual tools & $63.8$ & $65.6$ & $69.2$\\
    \hline
  \end{tabular}
  }
  \vskip -0.1in
  \caption{Different LLMs on the GQA and NLVRv2 datasets.}
  \label{tab:llmsgqa}
  \vskip -0.15in
\end{table}

\begin{table}
\centering
\resizebox{1\columnwidth}{!}{
\begin{tabular}{ c | c c c c}
    \toprule
       Method  & GQA & NLVRv2 & Editing & Tagging \\
    \hline
    LLama2-7B & $39.2 $ & $50.0 $ & {$31.2 $} & {$0.308$} \\
    LLama2-7B + Ours & $\mathbf{60.2 }$ & $\mathbf{63.8 }$ & {$\mathbf{47.6} $} & {$\mathbf{0.357}$} \\
    Mistral-7B & {$20.4 $} & {$34.6 $} & {$29.0 $} & {$0.205 $} \\
    Mistral-7B + Ours & {$\mathbf{31.4} $} & {$\mathbf{42.2} $} & {$\mathbf{46.5} $} & {$\mathbf{0.303}$} \\
    \bottomrule
\end{tabular}
}
  \vskip -0.1in
  \caption{Results on two open-source LLMs.}
  \vskip -0.15in
\label{table:opensource}
\end{table}

\section{Conclusion and Future Work}
\label{sec:conclusion}
In this paper, we have presented CLOVA, a general visual assistant that can adapt to new environments via inference, reflection, and learning in a closed-loop framework. In the inference phase, using both correct and incorrect examples for prompts benefits to generate better plans and programs. Our reflection scheme is capable of identifying tools that need to be updated. Through three data collection manners and the validation-learning prompt tuning scheme in the learning phase, CLOVA can efficiently improve its tools. Experimental results on four tasks and different LLMs show the effectiveness of CLOVA as a general visual assistant with learning abilities.

In the current method, we assume there is no selection or loop structure in programs, and assume there is at most one tool that needs updates in a task. The two assumptions cannot always hold in the real world. We could add selection and loop in-context examples for program generation and iterate the reflection and learning phases to update multiple tools. Besides, we could make some deployment designs to save the response time in our loop, including a foreground process and a background process. The former performs inference and gathers human feedback, while the latter does reflection, updates tools, and periodically synchronizes the weights of tools.

The framework of CLOVA can be easily generalized to new tools. (1) We will try multimodal LLMs (\emph{e.g.}, LLaVA-1.5~\cite{liu2023improved}) to replace LLMs. In this case, we will evaluate the effectiveness of visual information for plan and program generation, reflection, and answer inference. (2) We can add more up-to-date tools (\emph{e.g.}, BLIP2~\cite{li2023blip} as a VQA tool and SAM~\cite{Kirillov_2023_ICCV} as an SEG tool), by just designing and programming their forward and prompt tuning processes.

\noindent \textbf{Acknowledgements.}
We thank the anonymous reviewers for their constructive suggestions. Their insights have greatly improved the quality and clarity of our work. This work was supported in part by the National Science and Technology Major Project (2022ZD0114900).

\clearpage
\setcounter{page}{1}
\maketitlesupplementary



\section{Framework of CLOVA}
The pseudo-code of CLOVA is summarized in~\cref{alg:algorithm}.

\begin{algorithm}
\caption{CLOVA}
\label{alg:algorithm}
\small
\begin{algorithmic}[1] 
 \renewcommand{\algorithmicrequire}{\textbf{Input:}}
 \renewcommand{\algorithmicensure}{\textbf{Output:}}
 \REQUIRE LLMs, visual tools, instruction data $\mathcal{T}=\{T_1,T_2,\cdots,T_t \}$, demonstration pool $\mathcal{D}$, prompt pool $\mathcal{P}=\emptyset$.\\
\ENSURE Updated $\mathcal{D}$, updated $\mathcal{P}$ \\
\FOR {$i=1,2,\dots,t$}
\STATE Perform inference for $T_i$ by generating plan and program. 
\IF {$T_i$ is correctly solved}
\STATE Save the plan and program to $\mathcal{D}$.
\ELSE
\STATE Convert intermediate results into language.
\STATE Perform global reflection.
\STATE Perform training-validation prompt tuning, store in-context examples into $\mathcal{D}$ and prompts in $\mathcal{P}$.
\IF {Updated tools solve $T_i$ incorrectly}
\STATE Perform local reflection.
\STATE Perform training-validation prompt tuning, store in-context examples into $\mathcal{D}$ and prompts in $\mathcal{P}$.
\ENDIF
\ENDIF
\ENDFOR
\end{algorithmic}
\end{algorithm}

\section{Comparisons with related methods}
In~\cref{tab:related_work_supp}, we present a more detailed comparison table with more related methods. In the table, `Global' means the global reflection, `Local' means the local reflection, `Instruction' means instruction-following tuning, `RL' means reinforcement learning, and `Prompt' means using prompt examples as in-context learning. We observe that many tool-usage methods do not have a reflection capability, and few methods use global reflection to improve the plans or programs generated by LLMs. Different from them, CLOVA uses both global reflection and local reflection to identify tools that need to be updated, capable of handling complex instructions.
Moreover, as we all know, our method is the first work to update visual tools, through which the visual assistant can better adapt to new environments.

\begin{table*}
  \centering
  \begin{tabular}{c |c c c c}
    \toprule
    {Method} & Visual Tool & Reflection & Update LLMs & Update Tools\\
    \hline
    ART~\cite{Paranjape2023ARTAM} & \xmark  & \xmark & Prompt & - \\
    LATM~\cite{Cai2023LargeLM} & \xmark & Global & Prompt & - \\
    TRICE~\cite{Qiao2023MakingLM} & \xmark  & Global & Instruction + RL & - \\
    ToolkenGPT~\cite{hao2023toolkengpt} & \xmark  & - & \xmark & -  \\
    Toolformer~\cite{Schick2023ToolformerLM} & \xmark & - & Fine-tune & -  \\
    \hline
     VISPROG~\cite{gupta2023visual} &\cmark  & \xmark  & \xmark & \xmark \\
     Visual ChatGPT~\cite{wu2023visual} &\cmark & \xmark &\xmark & \xmark\\
     InternGPT~\cite{liu2023internchat} &\cmark  & \xmark  & \xmark & \xmark \\
     HuggingGPT~\cite{shen2023hugginggpt}  &\cmark & \xmark & \xmark & \xmark \\
     ViperGPT~\cite{surismenon2023vipergpt} &\cmark & \xmark & \xmark &\xmark\\
     ToT~\cite{hu2023tree} & \cmark & \xmark & \xmark & \xmark  \\
     Chameleon~\cite{lu2023chameleon} & \cmark & \xmark  & \xmark & \xmark \\
     ControlLLM~\cite{2023controlllm} & \cmark &  \xmark &\xmark & \xmark \\
     MM-REACT~\cite{yang2023mm} &\cmark & \xmark & \xmark & \xmark \\ 
     VideoAgent~\cite{fan2024videoagent} &\cmark & \xmark & \xmark & \xmark \\
     Llava-plus~\cite{liu2023llavaplus} &\cmark & \xmark & Instruction & \xmark\\
     Gorilla~\cite{patil2023gorilla} &\cmark & \xmark & Instruction & \xmark \\
     GPT4TOOLs~\cite{yang2023gpt4tools} &\cmark & \xmark & Instruction  & \xmark\\
     MLLM-Tool~\cite{wang2024mllmtool} &\cmark & \xmark & Instruction & \xmark\\
     VIoTGPT~\cite{zhong2023viotgpt} &\cmark & \xmark & Instruction & \xmark\\
     OpenAGI~\cite{ge2023openagi} &\cmark & \xmark & Reinforcement Learning & \xmark \\
     AssistGPT~\cite{gao2023assistgpt} &\cmark & Global & Prompt & \xmark\\
     \midrule
     \textbf{CLOVA (Ours)}&\cmark & Global+Local & Prompt & Prompt\\
    \bottomrule
  \end{tabular}
  \caption{Comparisons with representative tool-usage methods.
  }
  \label{tab:related_work_supp}
\end{table*}

\section{Prompt Examples}

\subsection{In-context examples}

\begin{figure*}
    \centering
    \includegraphics[width=1\textwidth]{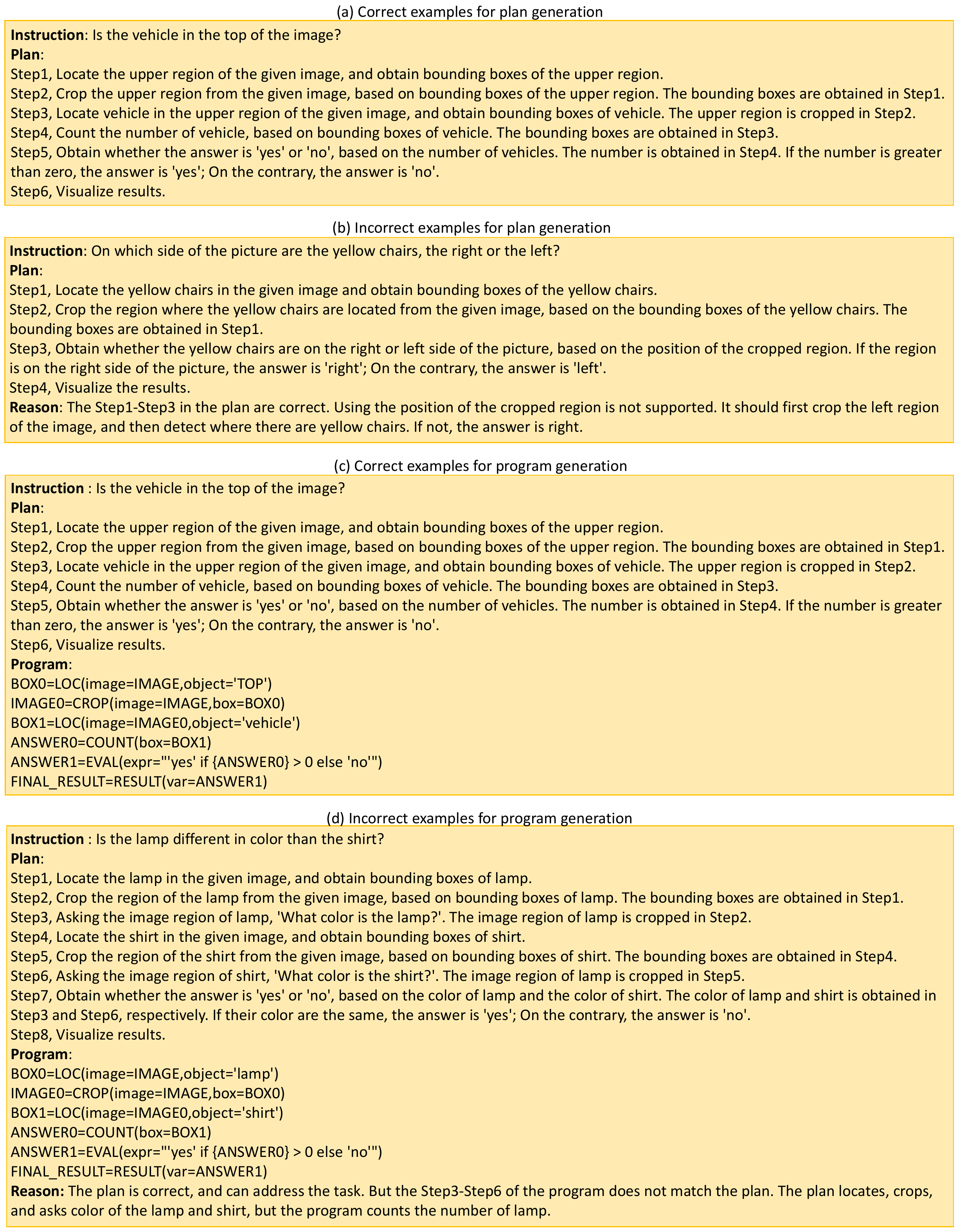}
      \caption{Demonstration examples for compositional VQA tasks in $\mathcal{D}_{p,s}$, $\mathcal{D}_{p,f}$, $\mathcal{D}_{c,s}$, and 
      $\mathcal{D}_{c,f}$. 
      }
\label{fig:demonstration_gqa}
\end{figure*}

\begin{figure*}
    \centering
    \includegraphics[width=0.85\textwidth]{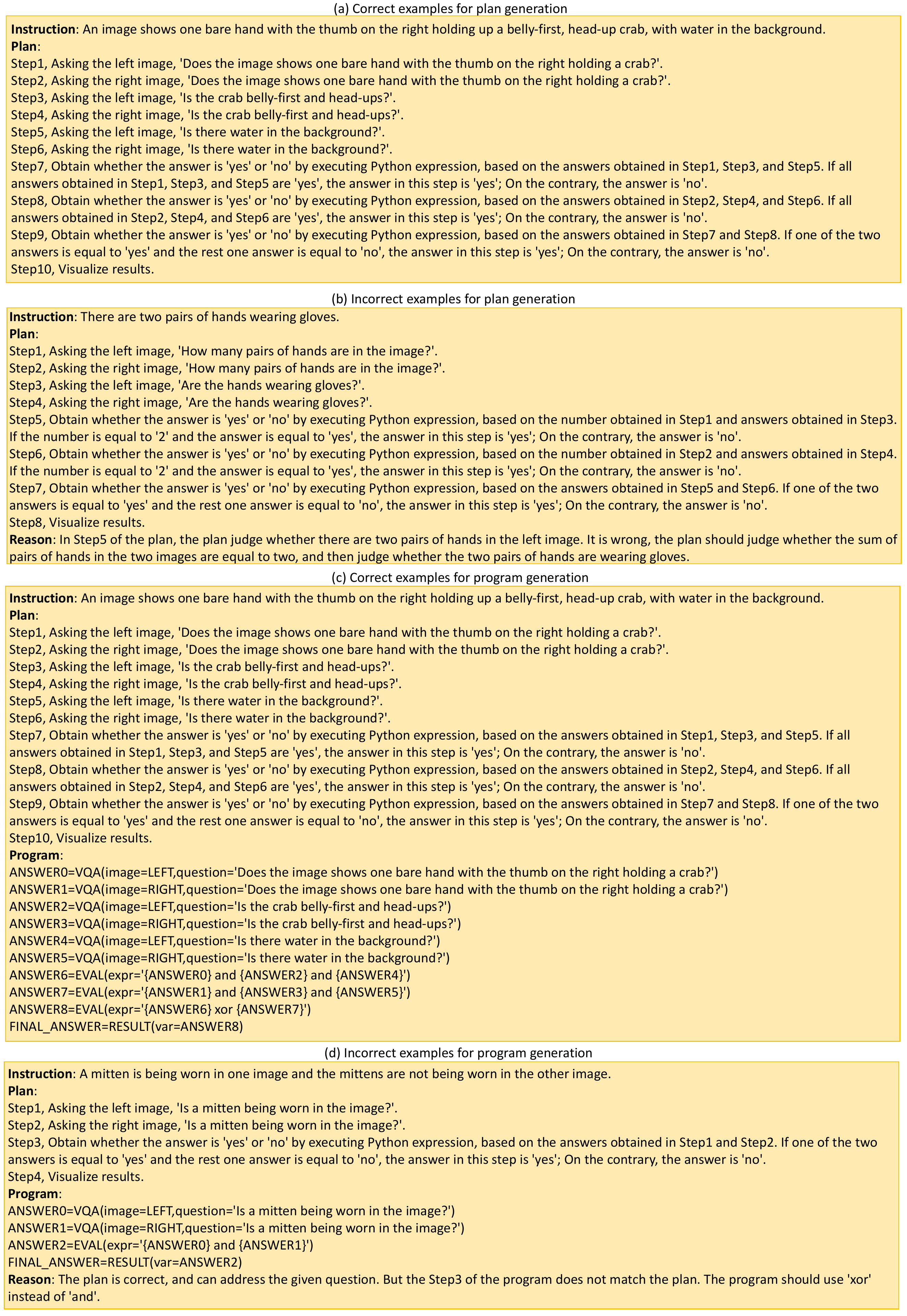}
      \caption{Demonstration examples for multi-image reasoning tasks in $\mathcal{D}_{p,s}$, $\mathcal{D}_{p,f}$, $\mathcal{D}_{c,s}$, and 
      $\mathcal{D}_{c,f}$. 
      }
\label{fig:demonstration_nlvrv2}
\end{figure*}

\begin{figure*}
    \centering
    \includegraphics[width=1\textwidth]{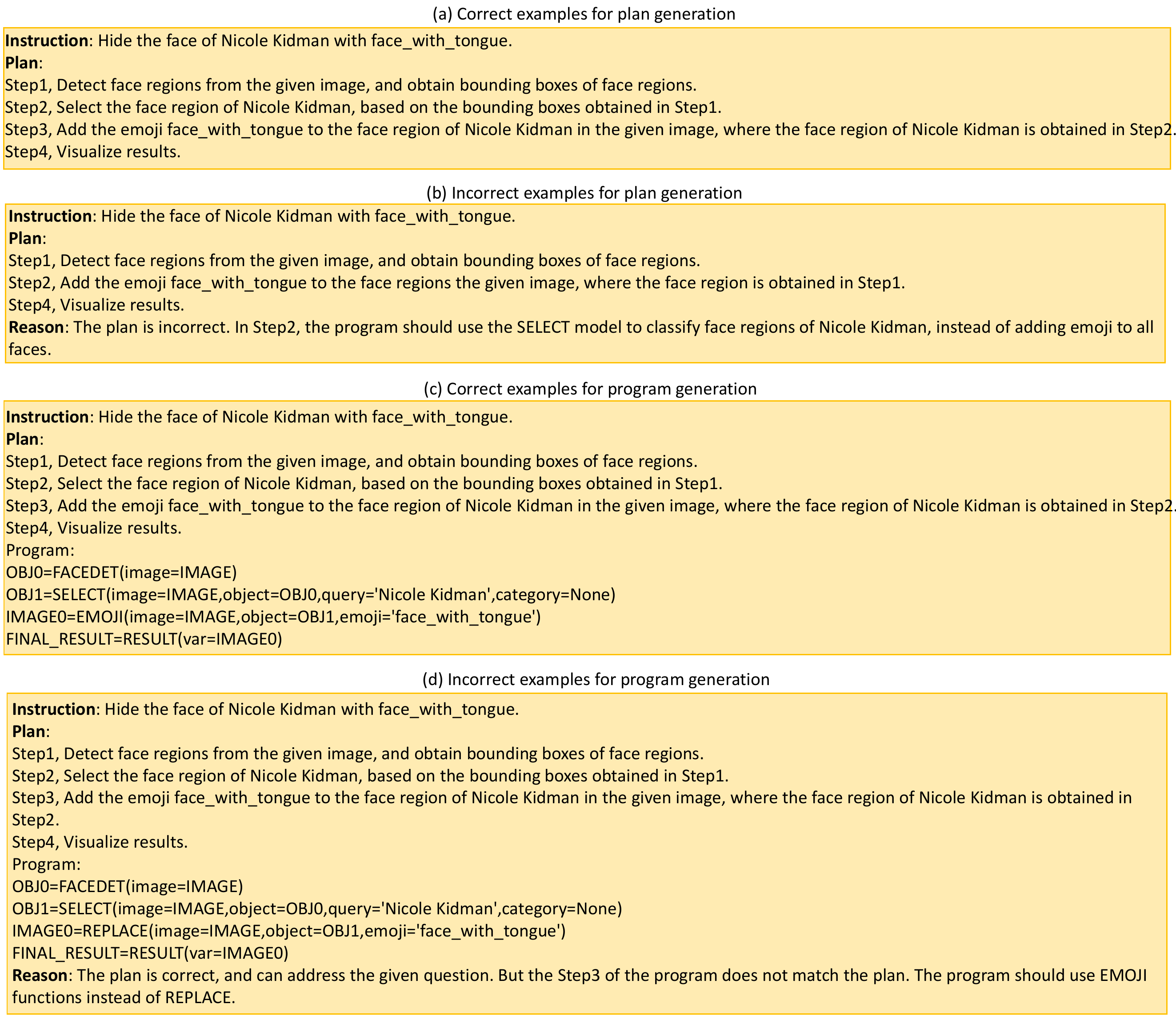}
      \caption{Demonstration examples for image editing tasks in $\mathcal{D}_{p,s}$, $\mathcal{D}_{p,f}$, $\mathcal{D}_{c,s}$, and 
      $\mathcal{D}_{c,f}$. 
      }
\label{fig:demonstration_editing}
\end{figure*}

\begin{figure*}
    \centering
    \includegraphics[width=1\textwidth]{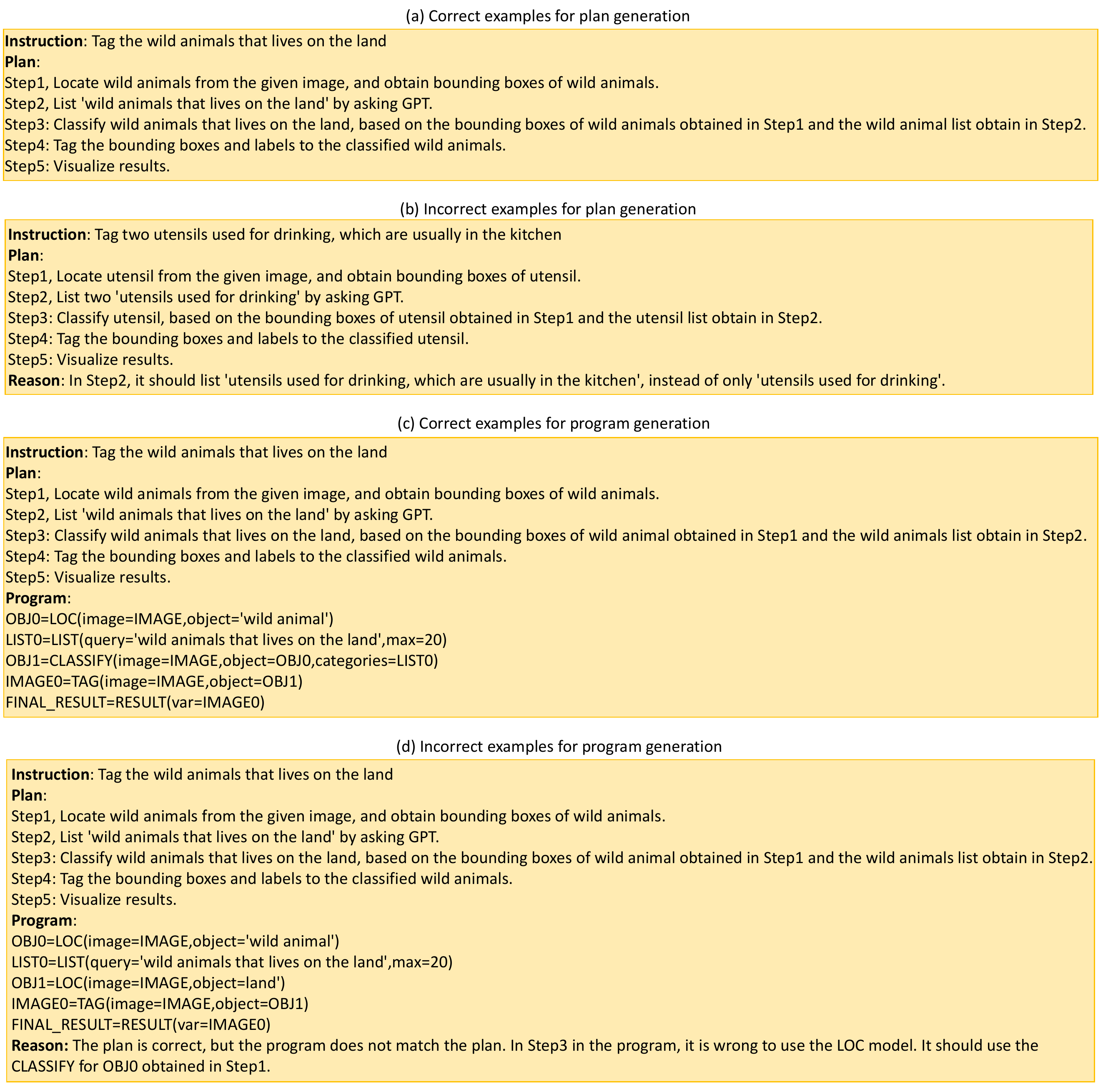}
      \caption{Demonstration examples for factual knowledge taggings task in $\mathcal{D}_{p,s}$, $\mathcal{D}_{p,f}$, $\mathcal{D}_{c,s}$, and 
      $\mathcal{D}_{c,f}$. 
      }
\label{fig:demonstration_tagging}
\end{figure*}

In the inference phase of our method, CLOVA generates plans and programs based on in-context examples that include correct examples and incorrect examples with criteria. Here we show some correct examples and incorrect examples for the compositional VQA, multi-image reasoning, image editing, and factual knowledge tagging tasks, as shown in~\cref{fig:demonstration_gqa,fig:demonstration_nlvrv2,fig:demonstration_editing,fig:demonstration_tagging}.

\subsection{Prompts in inference}

\begin{figure*}
    \centering
    \includegraphics[width=1\textwidth]{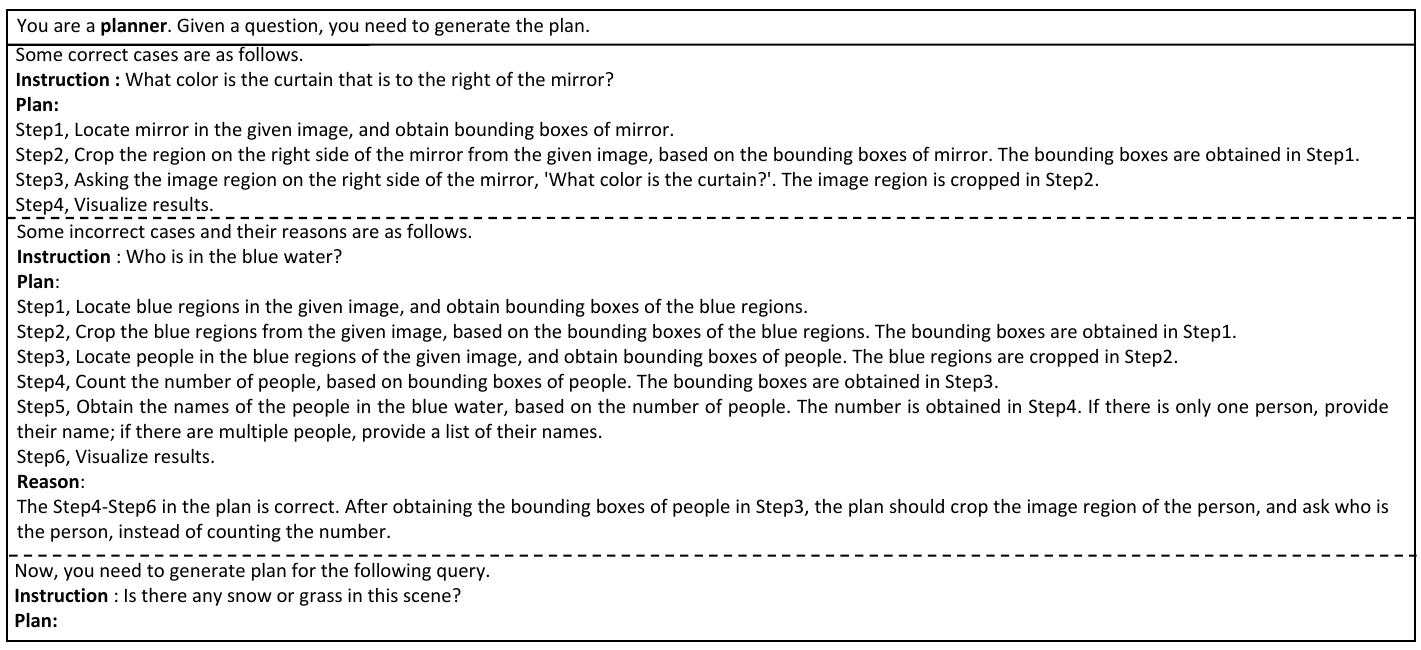}
      \caption{Prompts for plan generation.
      }
\label{fig:prompt_plan}
\end{figure*}

\begin{figure*}
    \centering
    \includegraphics[width=1\textwidth]{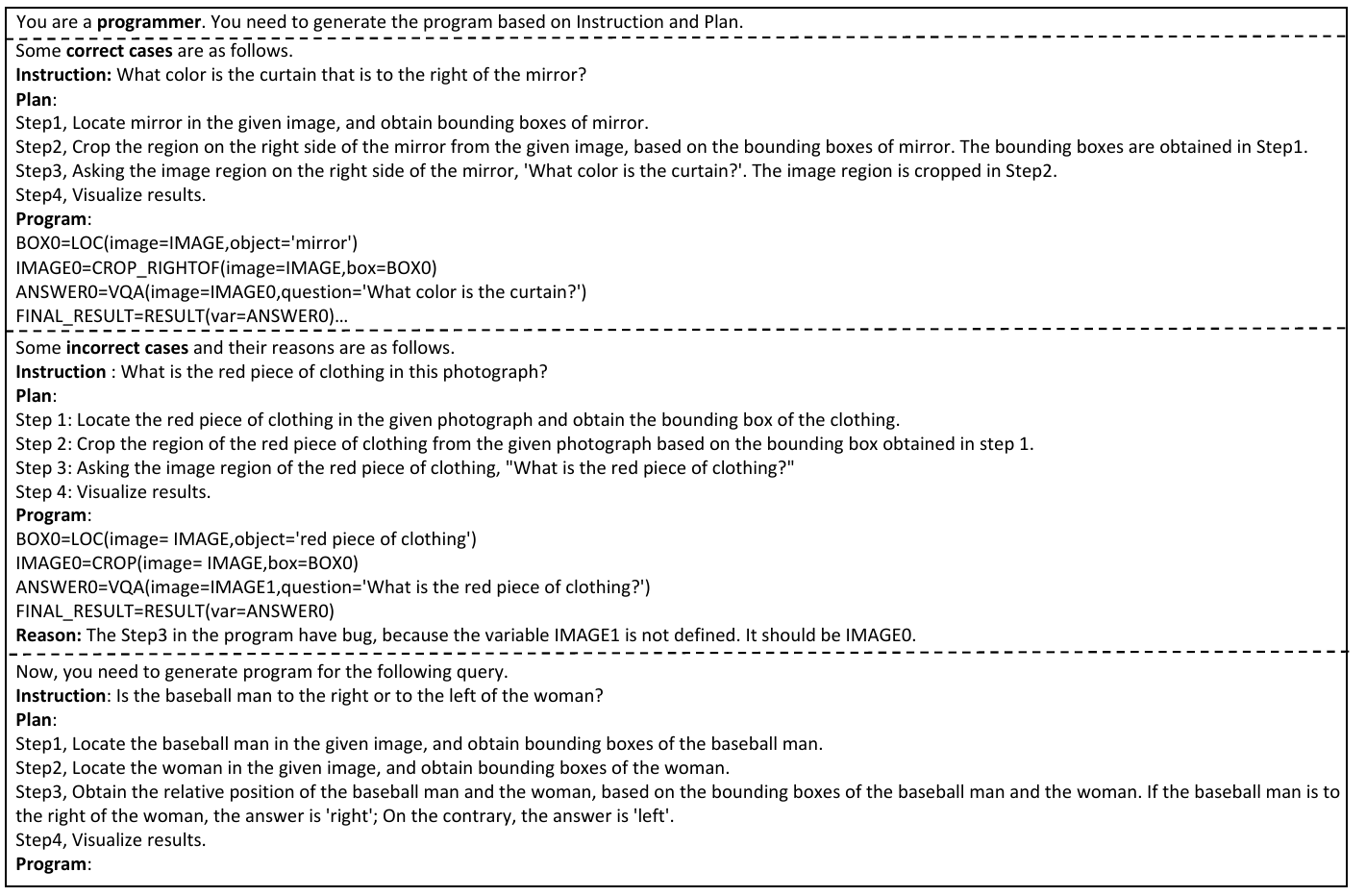}
      \caption{Prompts for program generation
      }
\label{fig:prompt_program}
\end{figure*}

In the inference phase, we use LLMs to generate plans and programs. We show examples of prompts for plan generation and program generation in~\cref{fig:prompt_plan,fig:prompt_program}, respectively.

\subsection{Prompts in reflection}
\begin{figure*}
    \centering
    \includegraphics[width=1\textwidth]{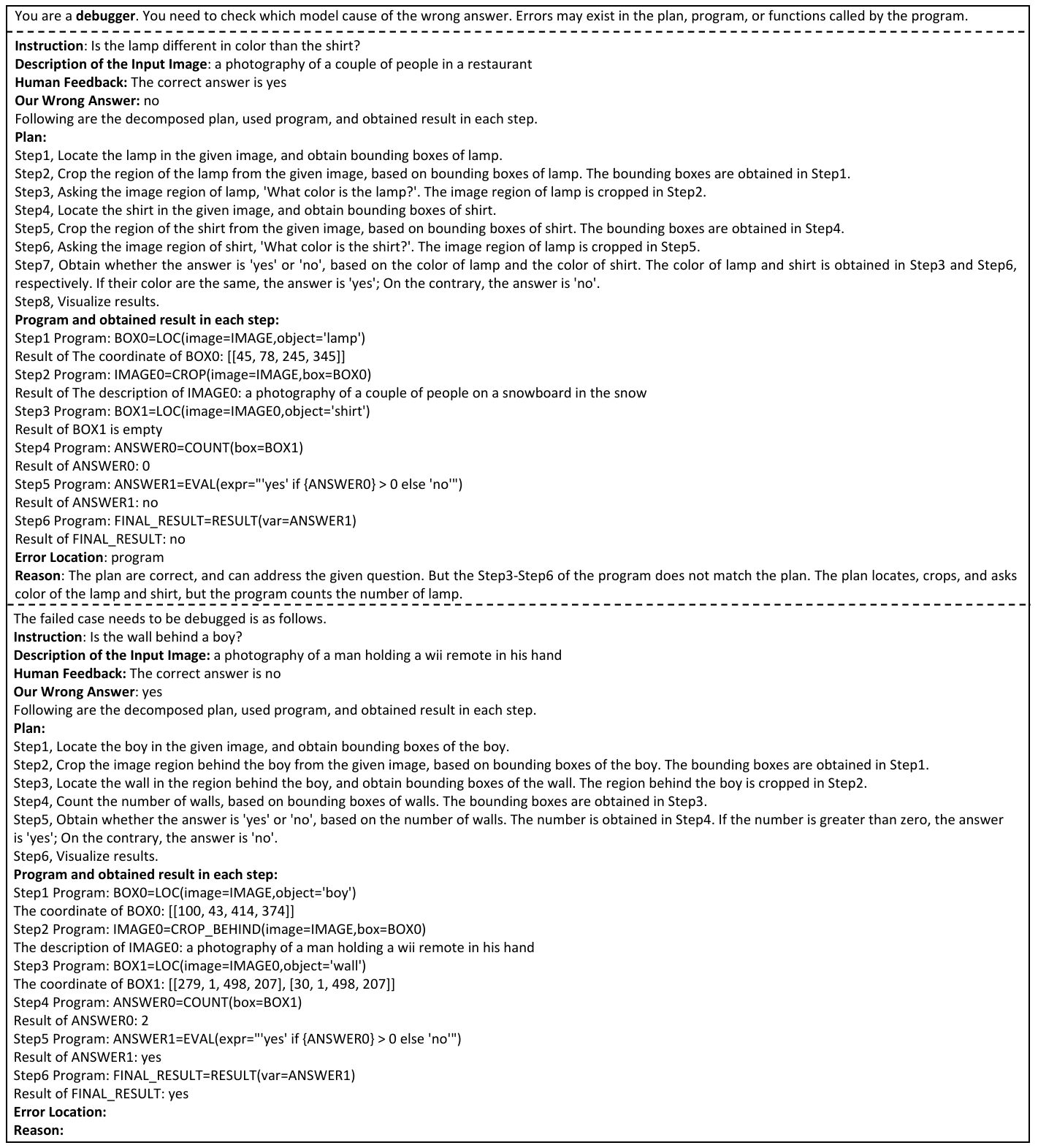}
      \caption{Prompts for global reflection.
      }
\label{fig:fast_reflection}
\end{figure*}

\begin{figure*}
    \centering
    \includegraphics[width=1\textwidth]{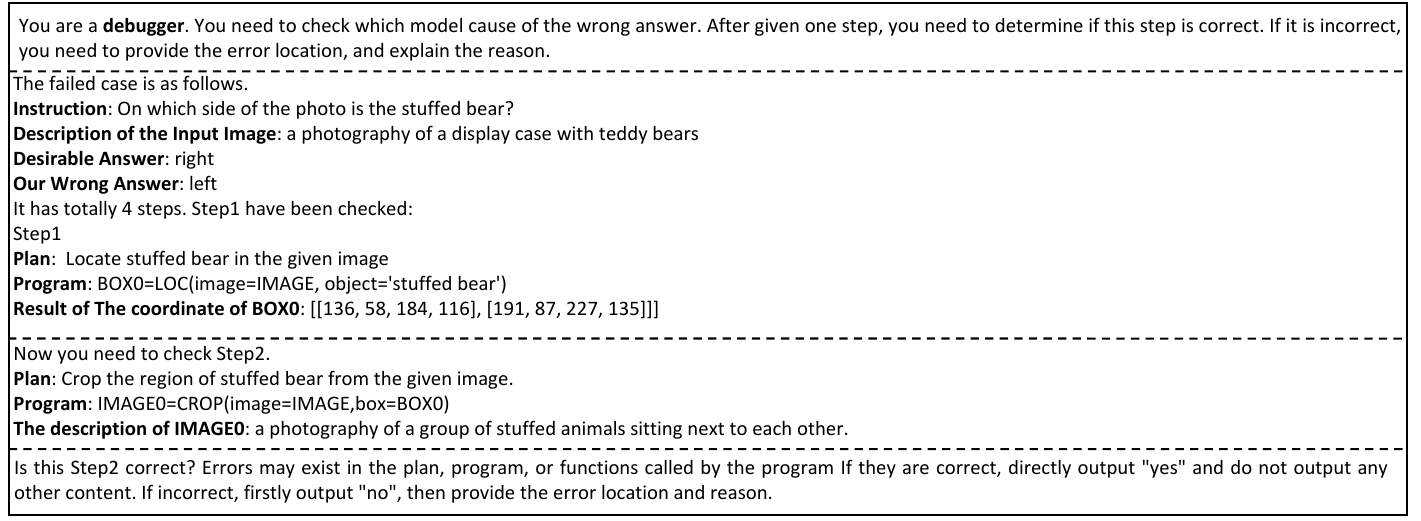}
      \caption{Prompts for local reflection.
      }
\label{fig:slow_reflection}
\end{figure*}

In the reflection phase, we use LLMs for global reflection and local reflection. We show two examples of prompts for global reflection and local reflection in~\cref{fig:fast_reflection,fig:slow_reflection}, respectively.

\subsection{Prompts in learning}
\begin{figure*}
    \centering
    \includegraphics[width=1\textwidth]{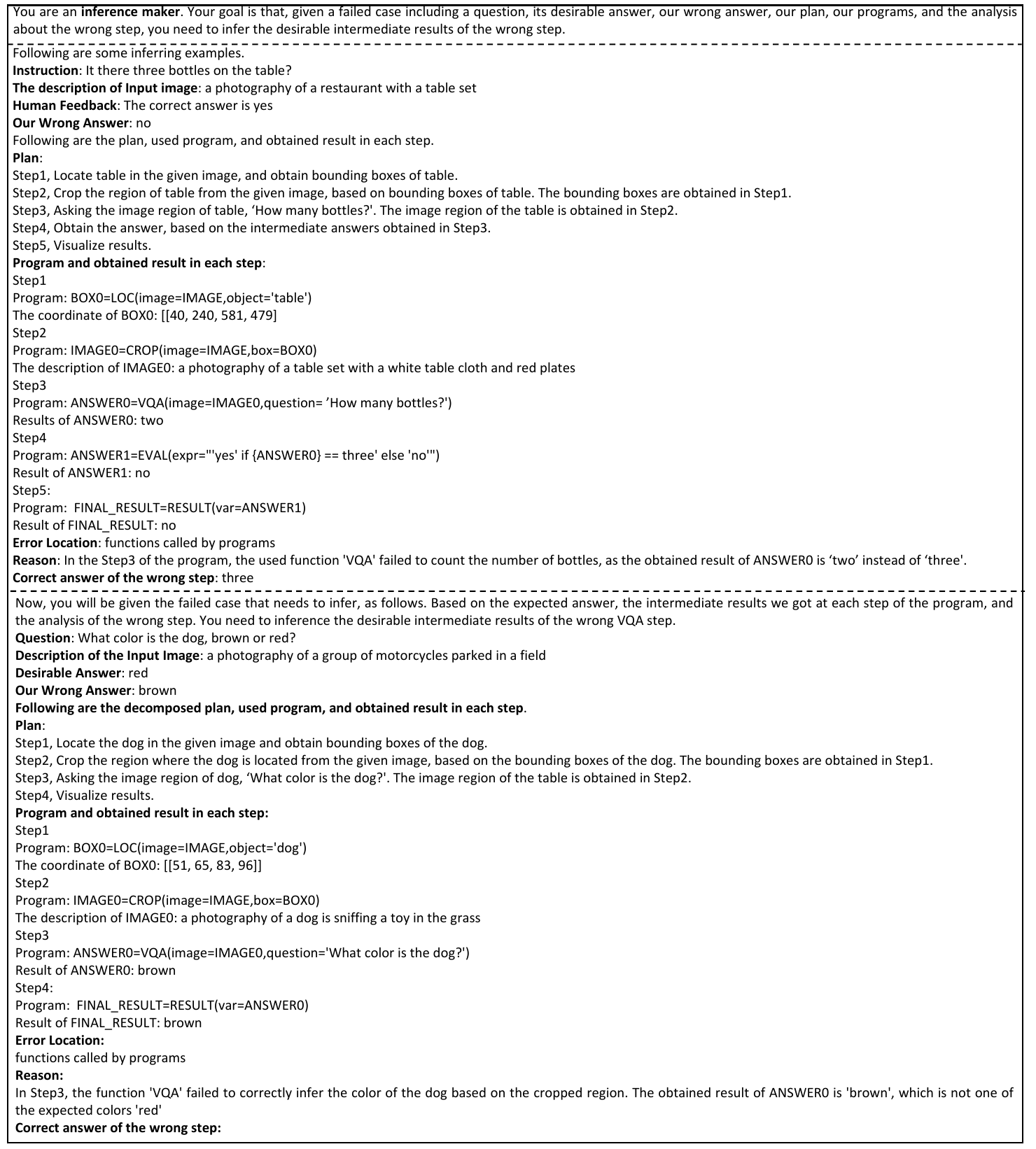}
      \caption{Prompts of inferring answers for the VQA tool.
      }
\label{fig:inference}
\end{figure*}

In the learning phase, we use LLMs to infer answers for the VQA tool, and then tune prompts of the VQA tool using the question and inferred answer. One example of prompts sent to LLMs for answer inferring is shown in~\cref{fig:inference}, respectively.

\section{Details of Tool update}

\subsection{Update VQA tool}

\subsubsection{Model}

\begin{figure*}
    \centering
    \includegraphics[width=1\textwidth]{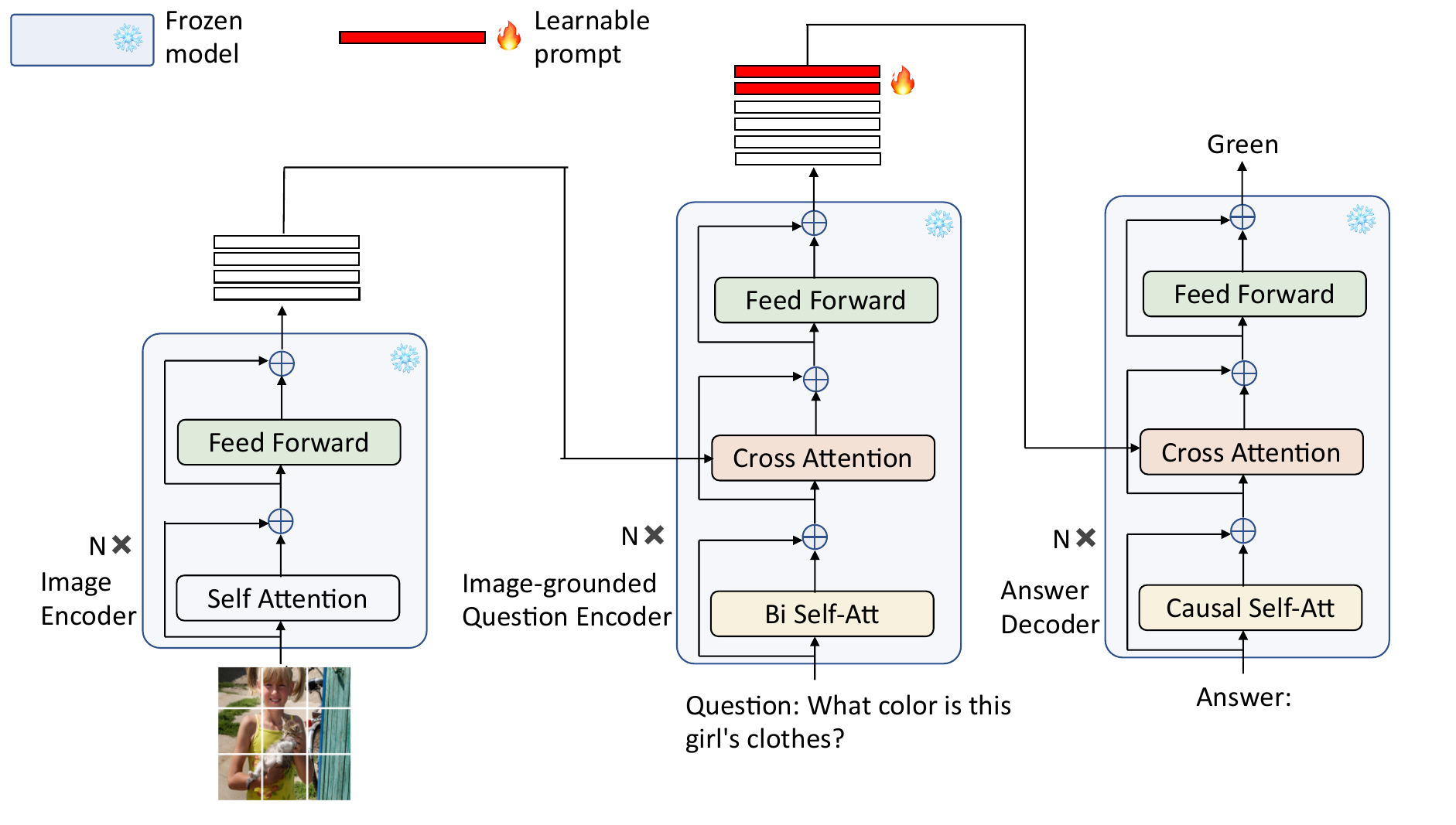}
      \caption{The architecture of BLIP.}
\label{fig:model_blips}
\end{figure*}

We use the BLIP~\cite{li2022blip} model for the VQA tool. One BLIP model is composed of three components: an image encoder, an image-grounded question encoder, and an answer coder, which are used to extract image features, extract question features, and generate answers, respectively.
In our prompt tuning scheme, the visual features are extracted from the image encoder followed by average pooling, whose dimension is $768$.
We concatenate learnable prompts with the inputs to the answer decoder, guiding the answer decoder to generate desirable answers. The architecture of BLIP is shown in Fig. \ref{fig:model_blips}.

\subsubsection{Training}
The size of learnable prompts is $\mathbb{R}^{16 \times 768}$, that is we learn $16$ vectors as prompts, and the dimension of each one is $768$. 
Since it is non-trivial to define visual concepts for the VQA tool, we do not store concepts in the prompt pool of the VQA tool, and all learned prompts and stored together.
We use questions and inferred answers of incorrect cases (detailed in Section 3.4) to update the VQA tool. We also store correct cases with zero vectors as prompts.
We use the language modeling loss~\cite{li2022blip} to train learnable prompts, where the Adam optimizer is used and the learning rate is $1e-3$. We train the prompts $100$ steps for each instance.

\subsubsection{Inference}
The prompt ensemble process of the VQA tool has two steps. (1) We roughly select out $20$ prompts from the prompt pool as candidates, by computing the similarity between the given query instance and stored instances in the prompt pool. (2) We use prompt ensemble (detailed in Section 3.4) to aggregate the $20$ prompts for a query instance.
In other words, we do not aggregate all stored prompts for a query instance, but $20$ similar instances.

\subsection{Update LOC tool}
\subsubsection{Model}
\begin{figure*}
    \centering
    \includegraphics[width=0.83\textwidth]{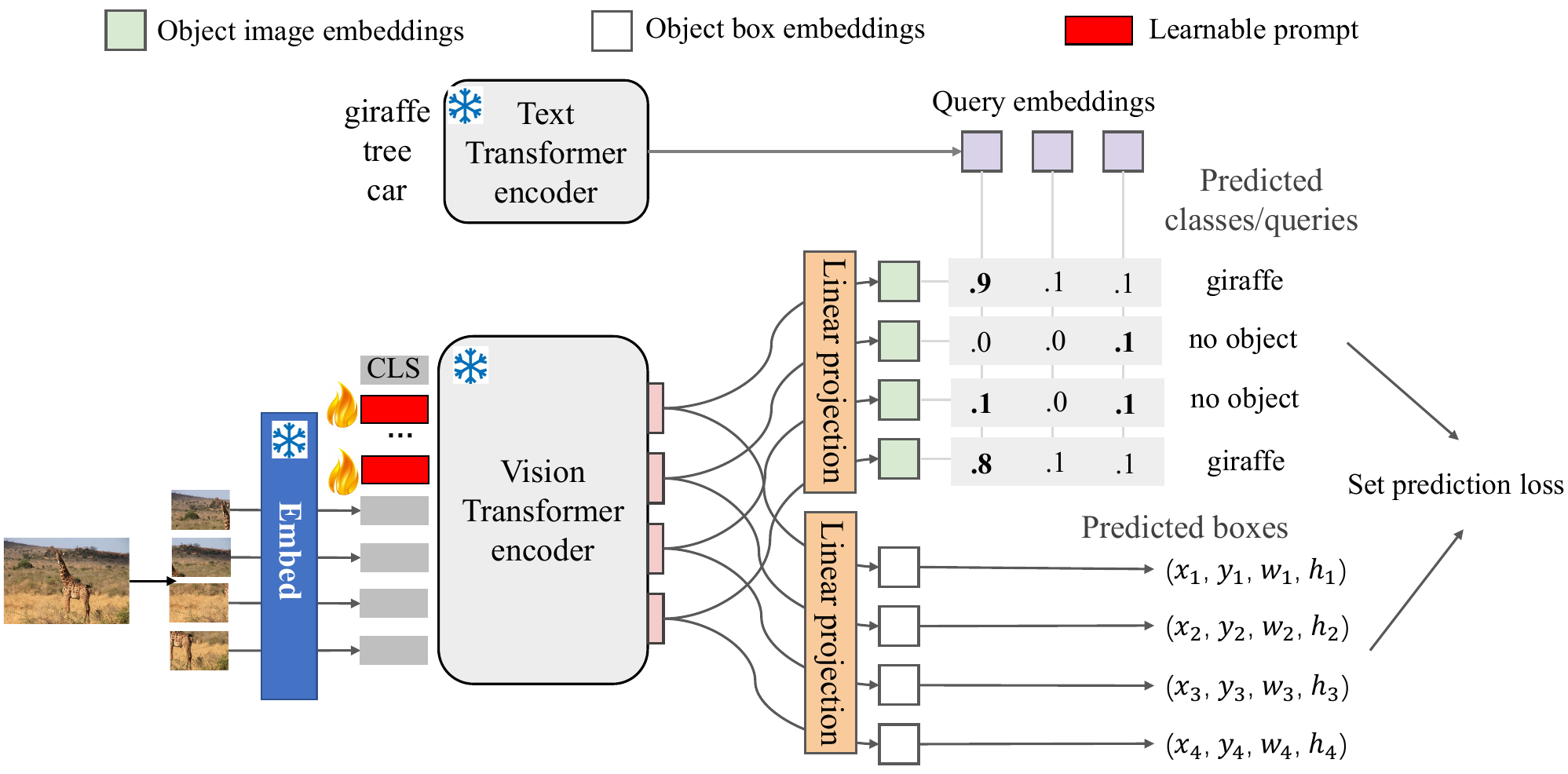}
      \caption{The architecture of OWL-ViT.}
\label{fig:model_owl_vit}
\end{figure*}
In CLOVA, we use the OWL-ViT model~\cite{LanguageModelscanSolveComputerTasks} for object localization as the LOC tool. The architecture of OWL-ViT is shown in Fig. \ref{fig:model_owl_vit}.
OWL-ViT model uses a standard vision transformer as the image encoder and a similar transformer architecture as the text encoder. It removes the token pooling and final projection layer, and instead linearly projects each output token representation
to obtain per-object image embeddings for classification. Besides, box coordinates are obtained by passing token representations through a small MLP.
The text embeddings, which are called queries, are obtained by passing category names or other textual object descriptions through the text encoder. 
At inference time, given a set of candidate class names and an image, the model predicts a bounding box and a probability
with which each query, and filters out the bounding box with the prediction confidence less than 0.1.

\subsubsection{Training}

In this study, upon identifying the need to update the LOC tool for learning a specific concept, we employ instance-wise prompts to update the OWL-ViT model.
To achieve this, CLOVA first collect training data from 
open-vocabulary datasets. We adopt LVIS dataset~\cite{gupta2019lvis} for OWL-ViT model. Taking the concept "glove" as an example, CLOVA randomly selects $200$ samples from the LVIS dataset, whose class labels contain a glove as the training data. During training, visual features are extracted from the backbone, 
and CLOVA concatenates learnable prompts with the inputs to the vision transformer decoder.
The model is trained with original losses introduced by OWL-ViT where only the prompts are learned. The losses include classification loss and bounding box regression loss. The former uses focal sigmoid cross-entropy~\cite{Zhu2020DeformableDD} while the latter uses $L_1$ loss. For classification loss, we regard the learned concept and one randomly selected class name in the image as positive labels and randomly select $13$ class names as negative labels per image to avoid overfitting. The number of learned prompts is $100$, and the prompt is randomly initialized. We set the maximum training step as $100$. If within $100$ steps, the sample could be detected correctly by the model, CLOVA would save this sample and the prompt, otherwise, CLOVA will remove this sample and the corresponding prompt. Only if all the positive labels are correctly classified and the average IOU between the prediction boxes and ground-truth boxes is larger than $0.6$, we recognize this sample is correctly detected. We use Adam as the optimizer and set the learning rate as $5e-4$. Eventually, we save the feature and the learned prompt of each correctly detected instance for this concept.

\subsubsection{Inference}

The prompt ensemble process of the LOC tool also has two steps. (1) Given a visual concept and a query image, we select all the prompts having the same visual concept from the prompt pool as candidates. We then filter out the prompts by making predictions with each candidate prompt, if the prediction confidence is larger than 0.1, we will use the prompt for the query image, otherwise, we will remove this candidate prompt. 
 (2) We use prompt ensemble (detailed in Section 3.4) to aggregate all the selected prompts for a query instance and contact the prompt with the input to produce a prediction.

\subsection{Update SEG tool}

\subsubsection{Model}

\begin{figure*}
    \centering
    \includegraphics[width=1\textwidth]{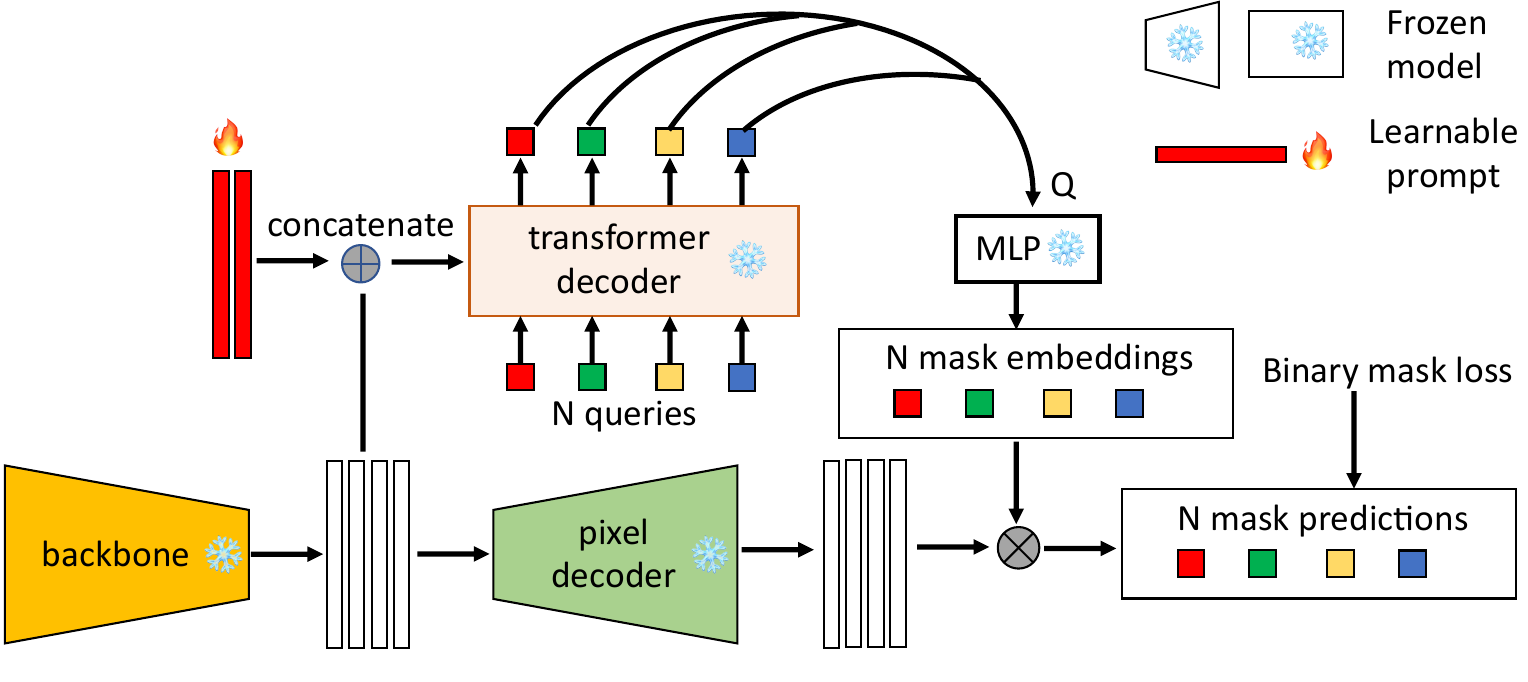}
      \caption{The architecture of Maskformer.}
\label{fig:model_maskformer}
\end{figure*}

We use the Maskformer~\cite{cheng2021per} model for the SEG tool. One Maskformer model is composed of three components: a backbone, a pixel decoder, and a transformer decoder. 
The backbone extracts features of images. Then, the pixel decoder gradually upsamples image features to extract per-pixel embeddings. Finally, the transformer decoder uses image features with our learnable prompts to generate per-mask embeddings that are combined with pre-pixel embedding for mask prediction. In our prompt tuning scheme, the visual features are extracted from the backbone followed by average pooling, whose dimension is $256$.
We concatenate learnable prompts with the inputs to the transformer decoder. 
The architecture of Maskformer is shown in Fig. \ref{fig:model_maskformer}.

\subsubsection{Training}
The size of learnable prompts is $\mathbb{R}^{100 \times 256}$, that is we learn $100$ vectors as prompts, and the dimension of each one is $256$. We remove the classification loss of Maskformer, and only use the mask loss~\cite{cheng2021per} to train learnable prompts, where the Adam optimizer is used and the learning rate is $1e-1$. 
Given a visual concept that needs to be learned, we randomly select $50$ samples having this visual concept from the LVIS dataset~\cite{gupta2019lvis}. We train the prompts $100$ steps for each instance.
After training, we store the concept name, and image features of instances, and learned prompts of image features in the the prompt pool.

\subsubsection{Inference}

The prompt ensemble process of the SEG tool also has two steps. (1)  Given a visual concept, we roughly select out $10$ prompts having the same visual concept from the prompt pool as candidates, by computing the similarity between the given query instance and stored instances in the prompt pool. (2) We use prompt ensemble (detailed in Section 3.4) to aggregate the $10$ prompts for a query instance.
In other words, we do not aggregate all stored prompts for a query instance, but $10$ similar instances.

\subsection{Update SELECT and  CLASSIFY tools}
\subsubsection{Model}
\begin{figure*}
    \centering
    \includegraphics[width=1\textwidth]{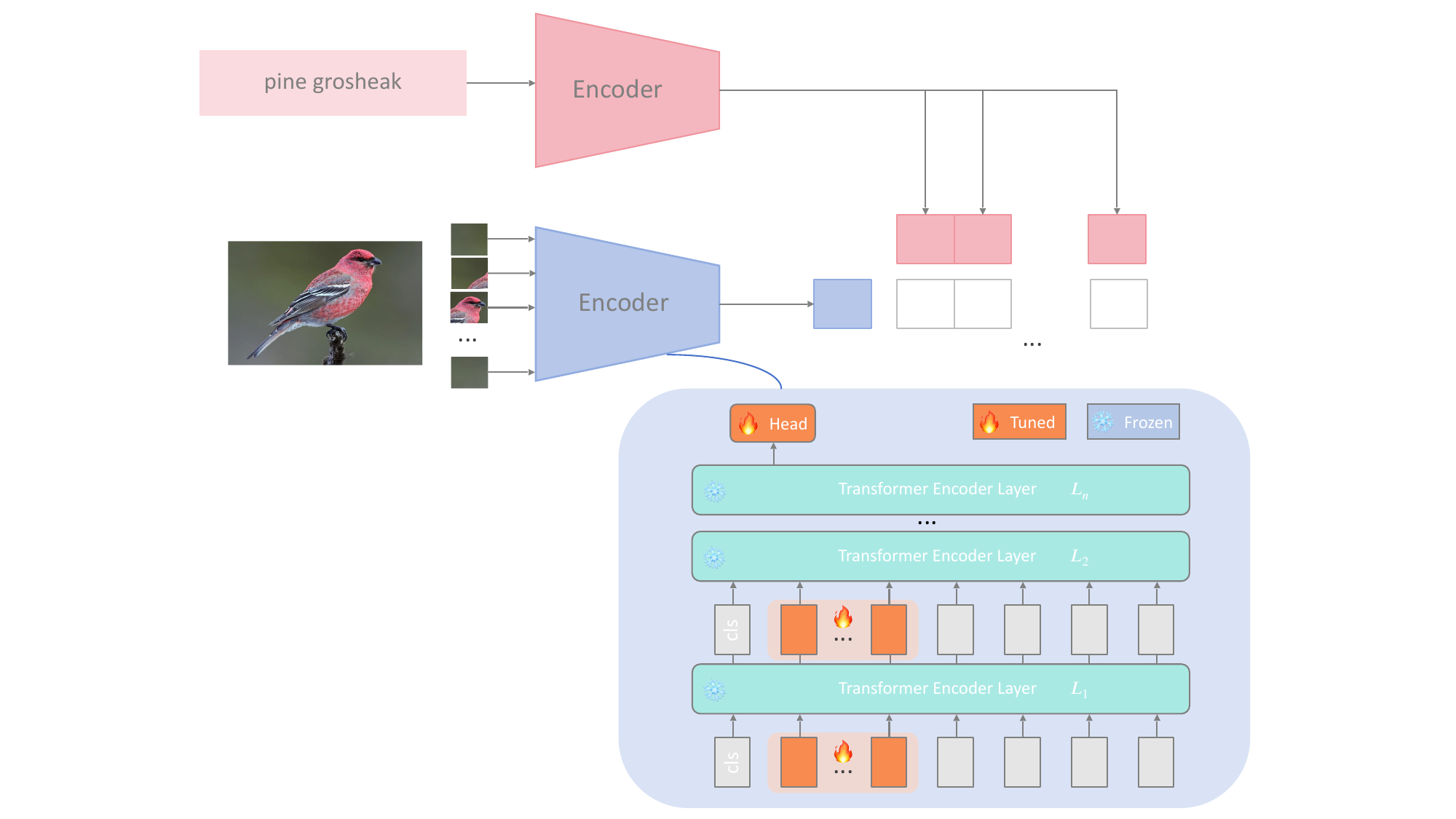}
        \caption{The architecture of CLIP.}
\label{fig:clip_update}
\end{figure*}

We utilize CLIP~\cite{radford2021learning} as the SELECT and CLASSIFY tools. The model consists of a Vision Transformer (ViT) as the image encoder and a Transformer-based text encoder. The image encoder and text encoder encode images and text descriptions into high-dimensional feature vectors respectively. It learns to align the representations of related content and separate unrelated content in the embedding space by utilizing a contrastive loss function. This loss function encourages CLIP to maximize the similarity between corresponding image-text pairs while minimizing it for non-corresponding pairs. 
In the prompt tuning scheme, learnable tokens are introduced into the image encoder and fine-tuned. These prompts are stored in the prompt pool for later use. During inference, the prompt is selected from the prompt pool and replaced with the image encoder to improve the precision of query data.
The architecture of CLIP is shown in Fig. \ref{fig:clip_update}.

\subsubsection{Training}
We update the CLIP through the deep prompt tuning~\cite{jia2022visual}. We utilize the Adam optimizer with a learning rate set to $1e-2$. To obtain training data, CLOVA uses the concept to be learned to automatically retrieve $7$ images from Google Images as positive data through web scraping. The first image is used for validation, while the remaining ones are used for training. 
Furthermore, CLOVA derives additional concepts related to but different from the target concept through GPT and subsequently collected data associated with these concepts as negative samples. CLOVA then uses both the positive data and negative data for training. 
During the training phase, we introduce $100$ learnable prompts for each of the first three layers of the vision transformer to facilitate prompt tuning. 
We conduct prompt tuning for $100$ steps per instance, followed by a validation step. If the accurate prediction is achieved on the validation data, CLOVA systematically stores features of crawled images and learned prompts in the prompt pool, otherwise, these learned prompts are discarded.

\subsubsection{Inference}
The prompt ensemble process of the SELECT and CLASSIFY tools also has two steps.
 (1) Given query data, we select data with its feature and prompt from the prompt pool with the same concept as the query data. Subsequently, we compute the similarity between the query data and the selected data. The prompts corresponding to data with high similarities are used for the query data. 
 (2) We use prompt ensemble (detailed in Section 3.4) to aggregate selected prompts for the query data.

\subsection{Update REPLACE tool}
\begin{figure*}
    \centering
    \includegraphics[width=1\textwidth]{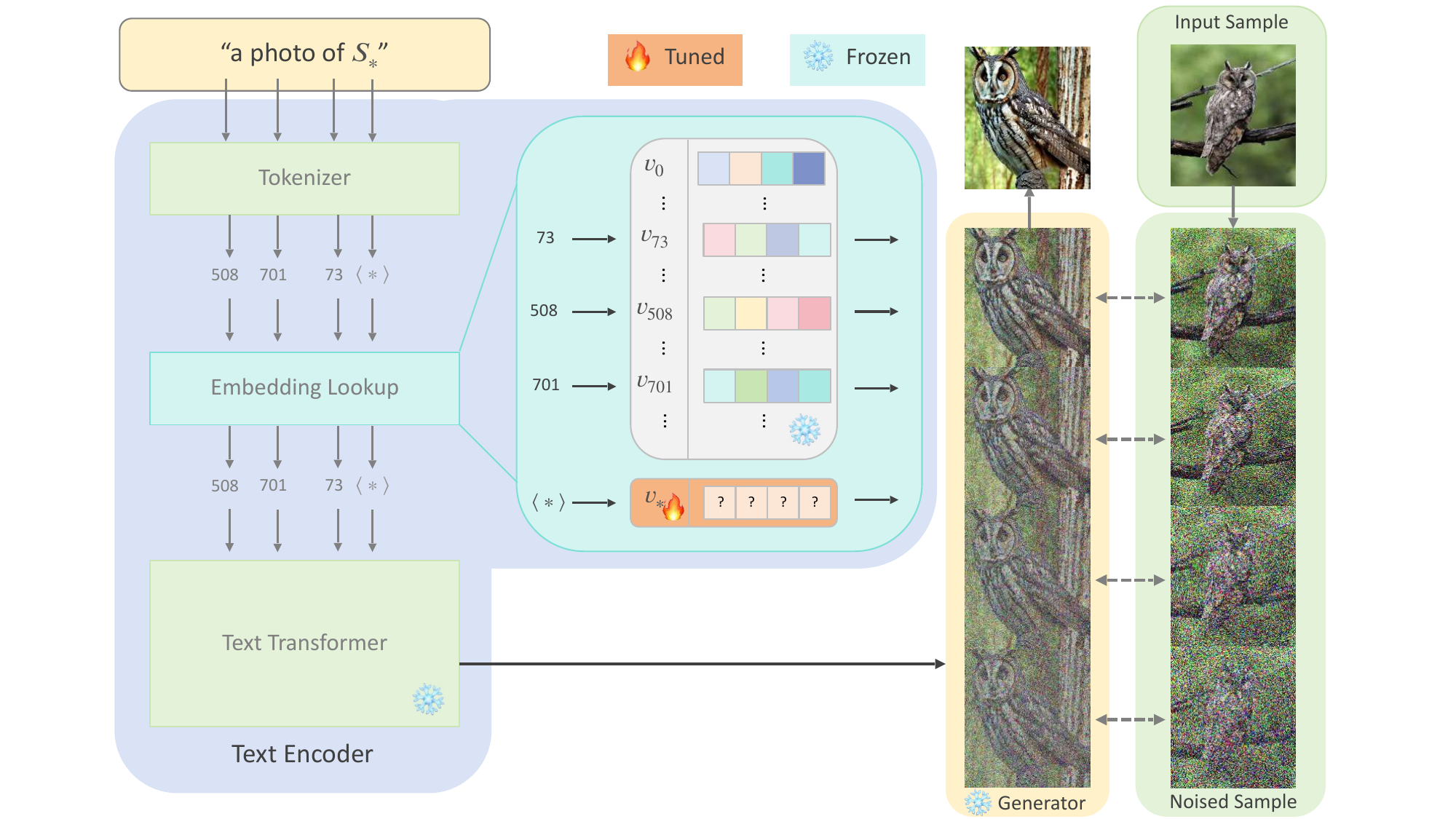}
        \caption{The architecture of Stable Diffusion.}
\label{fig:sd_update}
\end{figure*}

We use Stable Diffusion~\cite{rombach2022high} as the REPLACE tool. In Stable Diffusion, the architecture comprises four key components: Sampler, Variational Autoencoder (VAE), UNet, and CLIPEmbedder. The Sampler and UNet focus on the actual image generation, the VAE provides a deep understanding of image content, and the CLIPEmbedder ensures the relevance and accuracy of the generated images in relation to the text inputs. 
We added learnable prompts to the text encoder of the CLIPEmbedder component.
During the prompt tuning phase, we tune a prompt for training images and store it. The dimensionality of the prompt is $768$. 
The architecture of Stable Diffusion is shown in Fig. \ref{fig:sd_update}.

\subsubsection{Training}
In order to facilitate the learning of a specific concept for the Stable Diffusion model, CLOVA employs web scraping techniques to retrieve $7$ images representing this concept from Google Images. 
We train prompts in the text encoder of the Stable Diffusion model using downloaded images. During the training process, the Latent Diffusion Model(LDM) loss~\cite{rombach2022high} is minimized. Subsequently, the $768$-dimensional prompt obtained from the training stage is stored in the prompt pool. In terms of experimental setup, we employ the Adam optimizer. Through our experimentation, we find that setting the learning rate to $5e-3$ achieves the best learning results.

\subsubsection{Inference}
Similar to other visual tools, inference for the Stable Diffusion model also contains two steps. Given query data, CLOVA first locates prompts corresponding to the concept in the prompt pool and then loads the prompt into the text encoder of the Stable Diffusion model. Based on the query and mask, we perform editing on the input image.

\section{More Experimental Results}

\subsection{Training-validation prompt tuning for the VQA tool}
\begin{figure}
    \centering
    \includegraphics[width=0.48\textwidth]{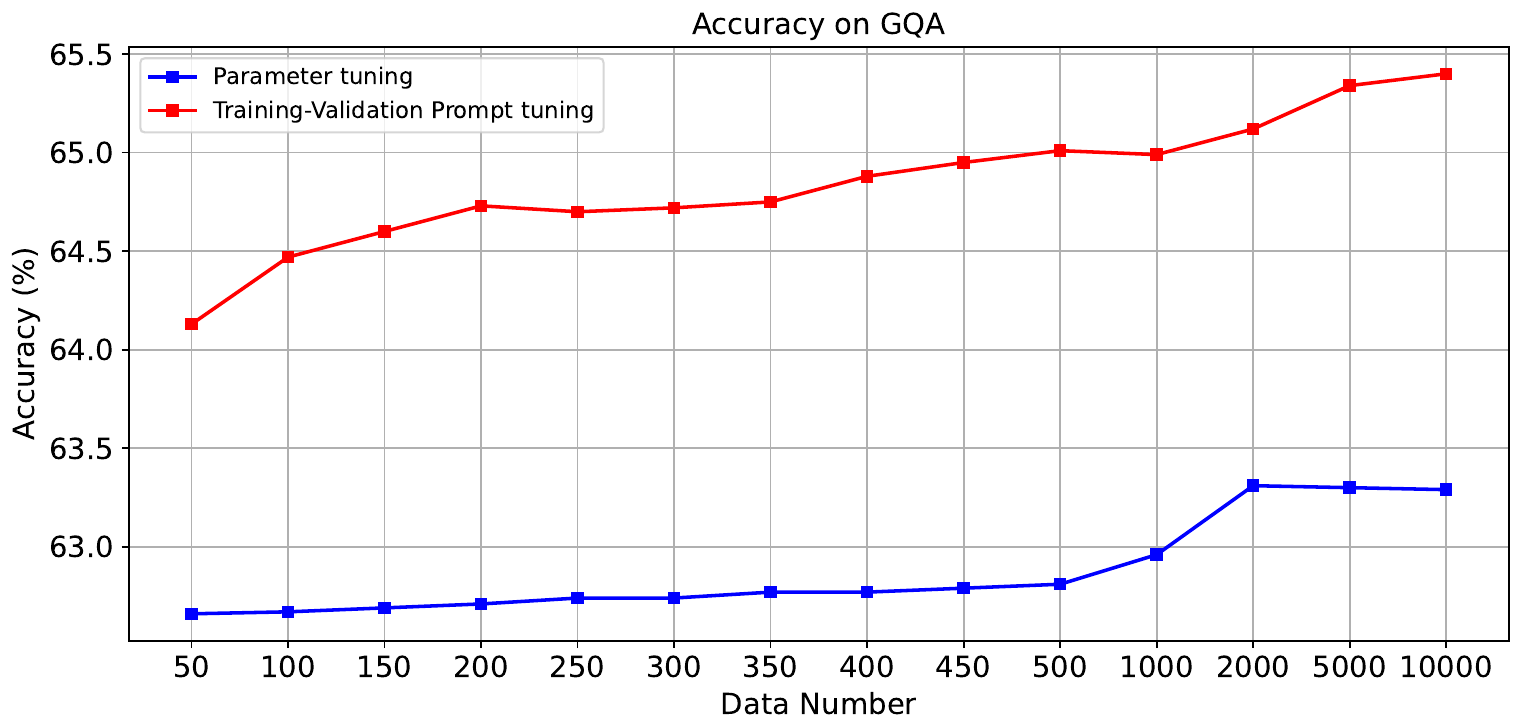}
      \caption{Accuracy curves on the GQA dataset}
\label{fig:Accuracy_on_GQA}
\end{figure}

We further evaluate the proposed validation-learning prompt tuning scheme for the VQA tool, where experiments are conducted on the compositional VQA task using the GQA dataset. 
We use the BLIP model for the GQA dataset and compare our prompt tuning scheme with direct tuning parameters. We report accuracies with using different numbers of training data. Results are shown in~\cref{fig:Accuracy_on_GQA}. Our method has higher performance throughout the entire training process, no matter whether the number of training data is small or large, showing the effectiveness of the proposed validation-learning prompt tuning scheme for the VQA tool.

\subsection{Evaluation on the online setting}

CLOVA can be applied to a more practical online learning setting. In this case, CLOVA is evaluated in a dynamic data stream. If it makes a correct prediction on a task, only the inference phase is activated, and the task is tagged as a correct prediction; if it makes an incorrect prediction on a task, this task is tagged as a wrong prediction, and the reflection and learning phases are activated to update tools. 
After the data stream, we calculate the accuracy based on tagged predictions of all cases.
We conduct experiments on the compositional VQA and multi-image reasoning tasks, where the GQA and NLVRv2 datasets are used. Results are shown in~\cref{tab:llmsgqaonline}. Similar to the offline setting in Section 4.2, updating LLMs and visual tools both leads to improvements.

\begin{table}
  \centering
  \resizebox{1\columnwidth}{!}{
  \begin{tabular}{c | c |c c c}
    \toprule
    Dataset  & Method & LLama2-7B & GPT-3.5-turbo & GPT-4 \\
    \hline
    \multirow{3}{*}{GQA} & Baseline & $39.2$ & $46.4$ & $52.6$\\
   &  + Update LLMs & $44.8$ & $51.0$ & $55.4$\\
   & + Update visual tools & $50.2$ & $53.0$ & $57.8$ \\      
    \hline
    \multirow{3}{*}{NLVRv2}  & Baseline & $50.0$ & $60.2$ & $64.8$\\
   & + Update LLMs & $57.4$ & $61.0$ & $66.4$\\
   & + Update visual tools & $61.6$ & $62.6$ & $67.4$\\
    \bottomrule
  \end{tabular}
  }
  \caption{Different LLMs on the online learning setting using the GQA and NLVRv2 datasets.}
  \label{tab:llmsgqaonline}
\end{table}

\subsection{More case studies}

\begin{figure*}
    \centering
    \includegraphics[width=1\textwidth]{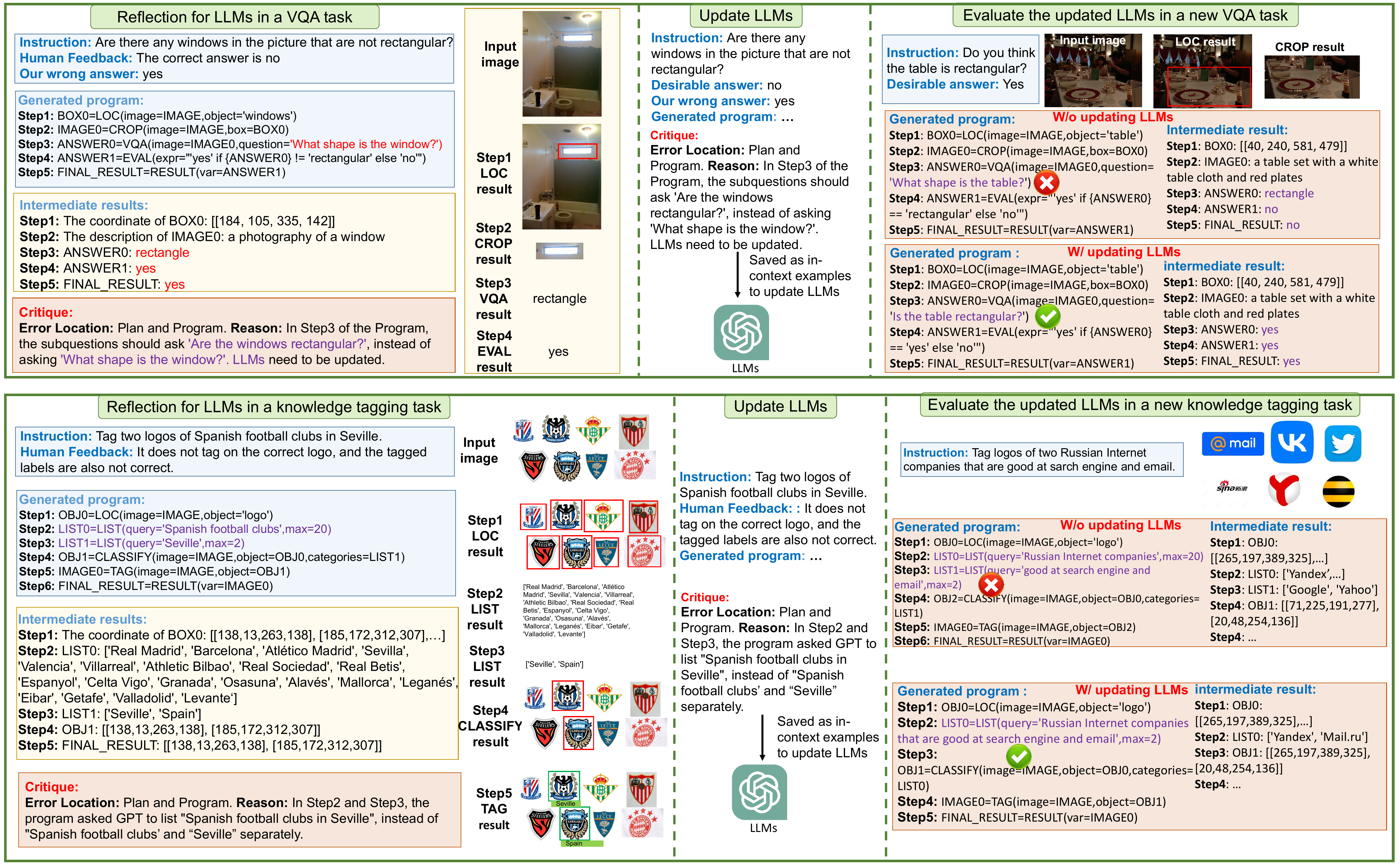}
      \caption{Case studies of updating LLMs.}
\label{fig:learning_LLMs}
\end{figure*}

\begin{figure*}
    \centering
    \includegraphics[width=1\textwidth]{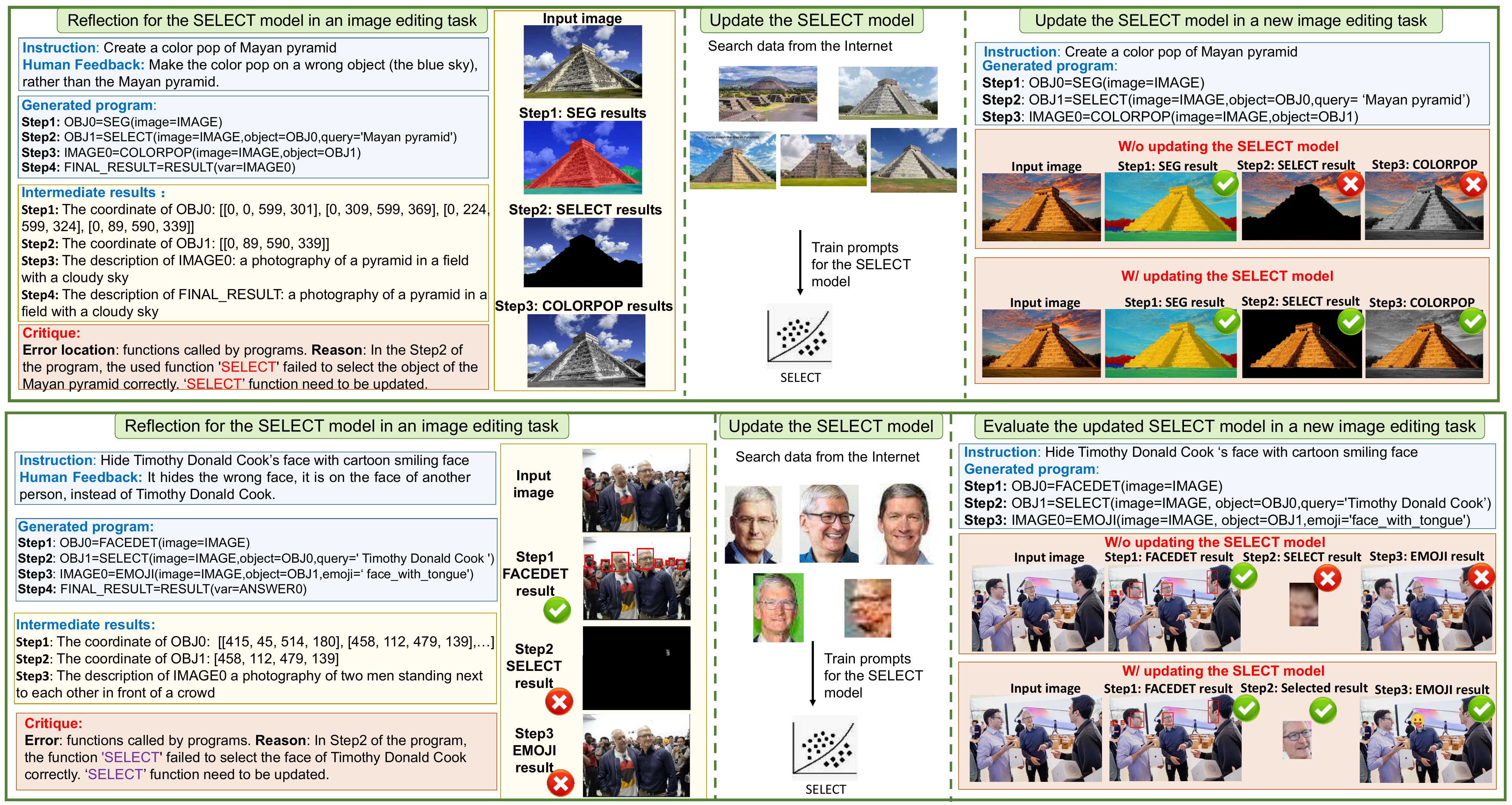}
      \caption{Case studies of updating the SELECT tool.}
\label{fig:learning_SELECT}
\end{figure*}

\begin{figure*}
    \centering
    \includegraphics[width=1\textwidth]{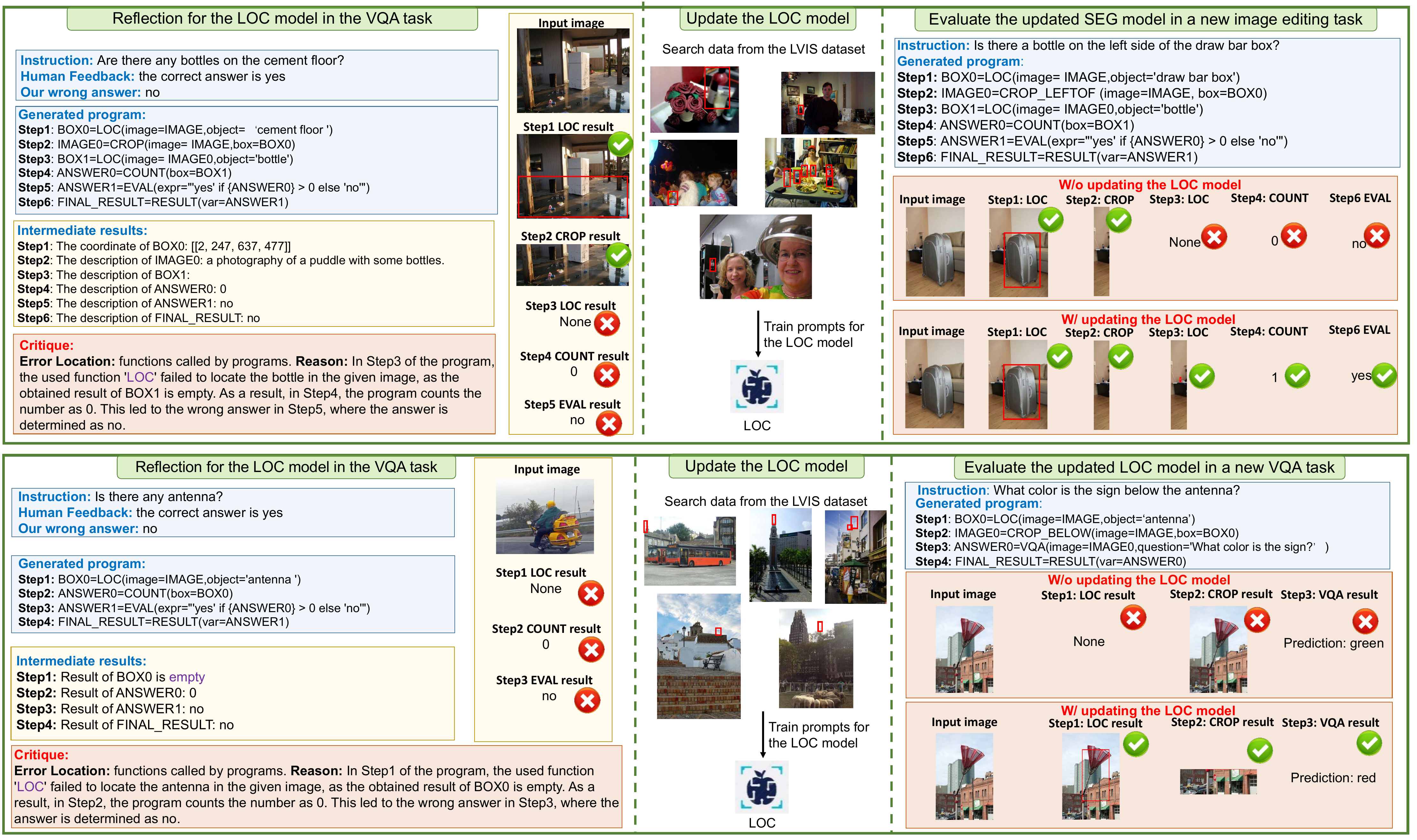}
      \caption{Case studies of updating the LOC tool.}
\label{fig:learning_LOC}
\end{figure*}

\begin{figure*}
    \centering
    \includegraphics[width=1\textwidth]{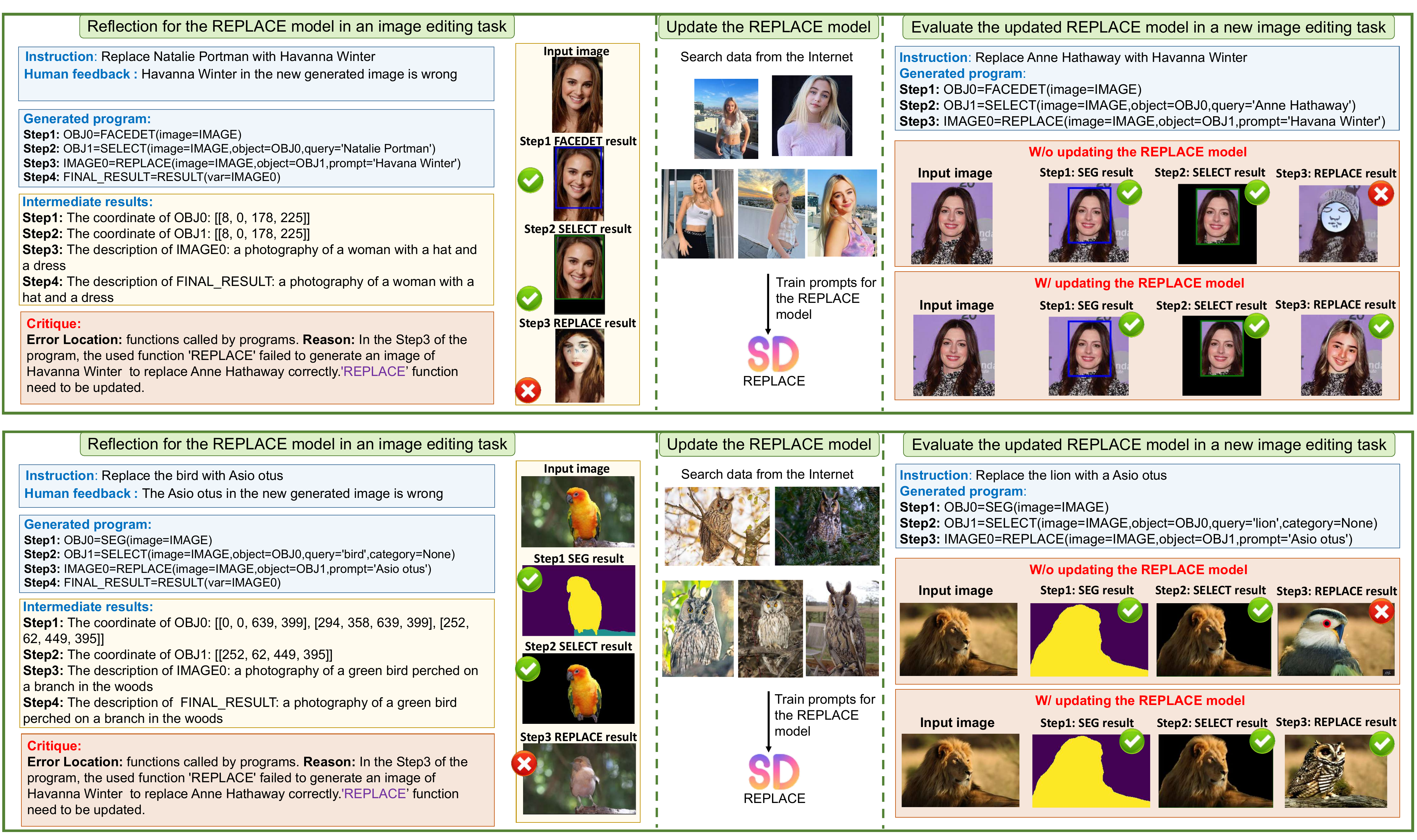}
      \caption{Case studies of updating the REPLACE tool.}
\label{fig:learning_REPLACE}
\end{figure*}

\begin{figure*}
    \centering
    \includegraphics[width=1\textwidth]{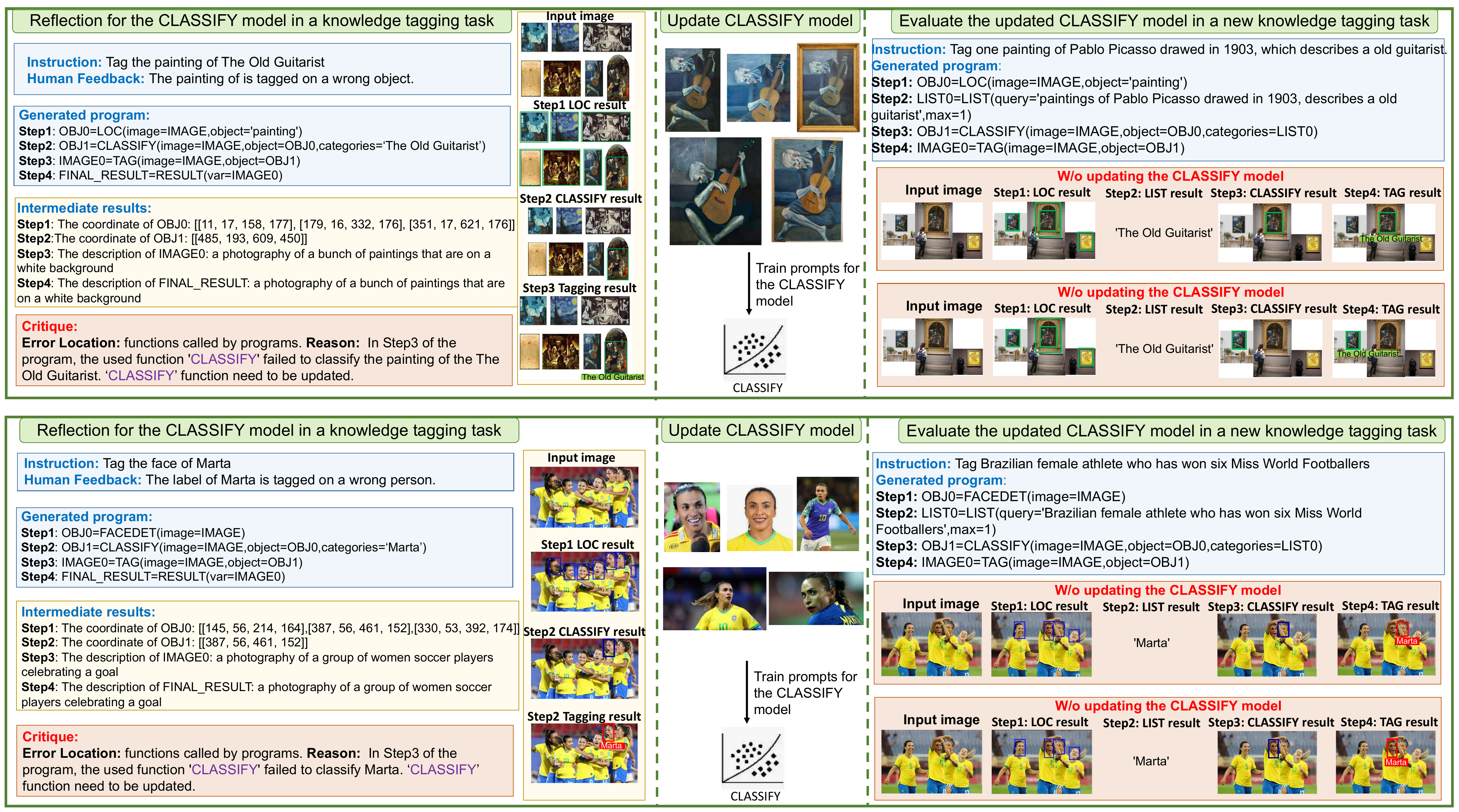}
      \caption{Case studies of updating the CLASSIFY tool.}
\label{fig:learning_CLASSIFY}
\end{figure*}

\begin{figure*}
    \centering
    \includegraphics[width=1\textwidth]{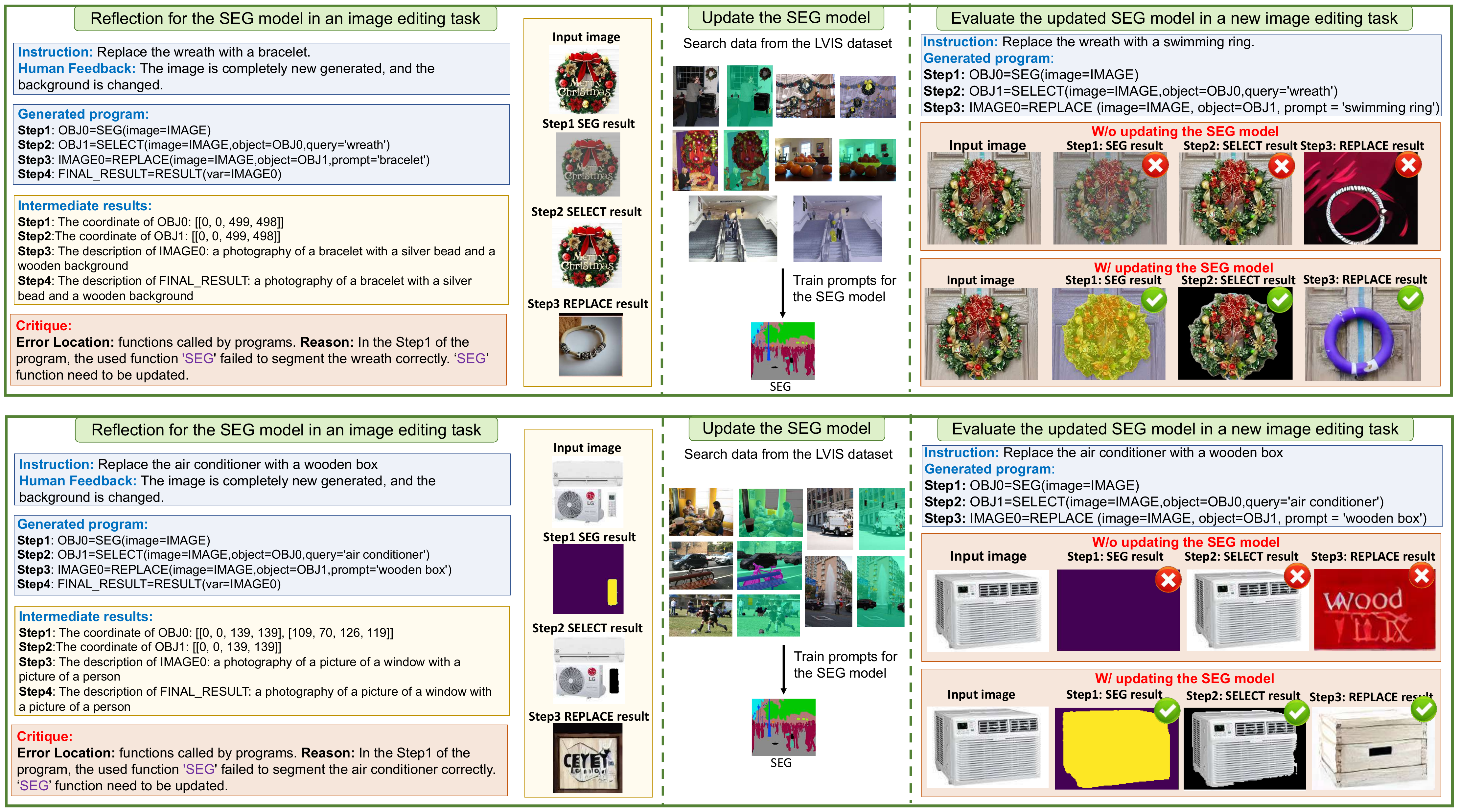}
      \caption{Case studies of updating the SEG tool.}
\label{fig:learning_SEG}
\end{figure*}

We provide more cases to show the reflection and learning phases of CLOVA.
The reflection and learning phases for LLMs are shown in~\cref{fig:learning_LLMs}.
The reflection and learning phases for the SELECT tool are shown in~\cref{fig:learning_SELECT}.
The reflection and learning phases for the LOC tool are shown in~\cref{fig:learning_LOC}.
The reflection and learning phases for the REPLACE tool are shown in~\cref{fig:learning_REPLACE}.
The reflection and learning phases for the CLASSIFY tool are shown in~\cref{fig:learning_CLASSIFY}.
The reflection and learning phases for the SEG tool are shown in~\cref{fig:learning_SEG}.

{
    \small
    \bibliographystyle{ieeenat_fullname}
    \bibliography{main}
}


\end{document}